\RequirePackage{fix-cm}
\documentclass[twocolumn]{svjour3}          
\smartqed  
\usepackage{graphicx}
\usepackage{epsfig}
\usepackage{amsmath}
\usepackage{amssymb}

\usepackage{xcolor}
\usepackage{soul}
\usepackage[utf8]{inputenc}
\usepackage[breaklinks=true,letterpaper=true,colorlinks,bookmarks=false,urlcolor=red,citecolor=cyan]{hyperref}

\usepackage{gensymb,graphics,subfigure,inputenc}
\usepackage[linesnumbered,algo2e,boxed]{algorithm2e}
\graphicspath{{Figures/}}
\usepackage[figuresright]{rotating}
\usepackage{amsmath,amssymb,textcomp,subfigure,multirow,upgreek}
\usepackage{booktabs}
\usepackage{color,mdwlist}

\usepackage{longtable}

\usepackage{ragged2e}

\usepackage{array}
\usepackage{tikz}
\usetikzlibrary{positioning, calc}
\definecolor{year2019}{RGB}{44, 123, 182}
\definecolor{year2020}{RGB}{102, 166, 30}
\definecolor{year2021}{RGB}{217, 95, 2}
\definecolor{year2022}{RGB}{117, 112, 179}
\definecolor{year2023}{RGB}{231, 41, 138}
\definecolor{timeline}{RGB}{50, 50, 50}
\usepackage[english]{babel}
\usepackage{microtype}

\usepackage{graphicx,fontawesome}
\graphicspath{{Figures/}}

\usepackage{amssymb} 
\usepackage{xcolor} 
\usepackage{caption}
\usepackage{wrapfig}
\graphicspath{ {image/} }  
\usepackage{enumitem}
\setitemize{noitemsep,topsep=0pt,parsep=0pt,partopsep=0pt}

\usepackage{tikz}
\usepackage{tikz-3dplot}
\usetikzlibrary{arrows.meta, positioning}

\tikzset{
    node distance=2cm and 1cm,
    every node/.style={fill=white, font=\sffamily},
    myarrow/.style={-Stealth, thick}
}

\definecolor{myblue}{HTML}{8FAADC}
\definecolor{myorange}{HTML}{F4B183}
\definecolor{mygray}{HTML}{DBDBDB}
\definecolor{myyellow}{HTML}{FFD966}
\definecolor{mypeach}{HTML}{CFA9C9}

\begin{document}

\title{Multimodal Alignment and Fusion: A Survey}


\author{       
Songtao Li$^*$ \and Hao Tang$^\dagger$  
}


\institute{
Songtao Li ($^*$work done during the visit at Peking University) \at Sydney Smart Technology College, Northeastern University, Qinhuangdao 066006, China. \email{202219226@stu.neu.edu.cn} \\
Hao Tang ($^\dagger$corresponding author) \at 
The State Key Laboratory of Multimedia Information Processing,
School of Computer Science, Peking University, Beijing 100871, China. 
\email{haotang@pku.edu.cn} 
}

\date{Received: date / Accepted: date}

\maketitle

\begin{abstract}
This survey provides a comprehensive overview of recent advances in multimodal alignment and fusion within the field of machine learning, driven by the increasing availability and diversity of data modalities such as text, images, audio, and video. Unlike previous surveys that often focus on specific modalities or limited fusion strategies, our work presents a structure-centric and method-driven framework that emphasizes generalizable techniques. We systematically categorize and analyze key approaches to alignment and fusion through both structural perspectives---data-level, feature-level, and output-level fusion---and methodological paradigms---including statistical, kernel-based, graphical, generative, contrastive, attention-based, and large language model (LLM)-based methods, drawing insights from an extensive review of over 260 relevant studies. Furthermore, this survey highlights critical challenges such as cross-modal misalignment, computational bottlenecks, data quality issues, and the modality gap, along with recent efforts to address them. Applications ranging from social media analysis and medical imaging to emotion recognition and embodied AI are explored to illustrate the real-world impact of robust multimodal systems. The insights provided aim to guide future research toward optimizing multimodal learning systems for improved scalability, robustness, and generalizability across diverse domains.


\end{abstract}

\section{Introduction}

Rapid advancement in technology has led to an exponential increase in the generation of multimodal data, including images, text, audio, and video \cite{tadas2019multimodal}. This abundance of data presents opportunities and challenges for researchers and practitioners in diverse fields, such as computer vision and natural language processing. Integrating information from multiple modalities can significantly enhance the performance of machine learning models, improving their ability to understand complex scenarios in the real world \cite{boehm2021harnessing,sdseg3d_eccv2022,mseg3d_cvpr2023,zhang2024risurconvrotationinvariantsurface,bachmann20244m21anytoanyvisionmodel,duan2021audio}.

At the core of multimodal learning lie two interdependent problems: \emph{alignment} and \emph{fusion}. Alignment aims to establish semantic correspondences across modalities so that their representations occupy a shared space, while fusion merges these aligned features into unified predictions or embeddings. The combination of modalities is generally pursued with two main objectives: (i) Different data modalities can complement each other, thus improving the precision and effectiveness of models for specific tasks \cite{li2024dataprocessingtechniquesmodern,zhang_deep_fusion_2021,ma_learning_2024}. (ii) Some modalities may have limited data availability or may be challenging to collect in large quantities; therefore, training in an LLM-based model can leverage knowledge transfer to achieve satisfactory performance in tasks with sparse data \cite{Liang2024Foundations,ma_learning_2024}.

For example, in social media analysis, combining textual content with related images or videos offers a more comprehensive understanding of user sentiment and behavior \cite{tadas2019multimodal,liu2023visualinstructiontuning}. Beyond social networks, multimodal methods have shown promising results in applications such as automated caption generation for medical images, video summarization, and emotion recognition \cite{gabeur_multi-modal_2020,mahdum2017applications,chen2023llavamed,yang2023videochat,zhu2023minigpt4enhancingvisionlanguageunderstanding,xie2025tttfusion}. Despite these advancements, two major technical challenges remain in effectively leveraging multimodal data: alignment—ensuring semantic consistency across modalities—and fusion—integrating complementary cues to enhance downstream performance.

\begin{figure*}[!t]
    \centering
    \includegraphics[width=0.8\textwidth]{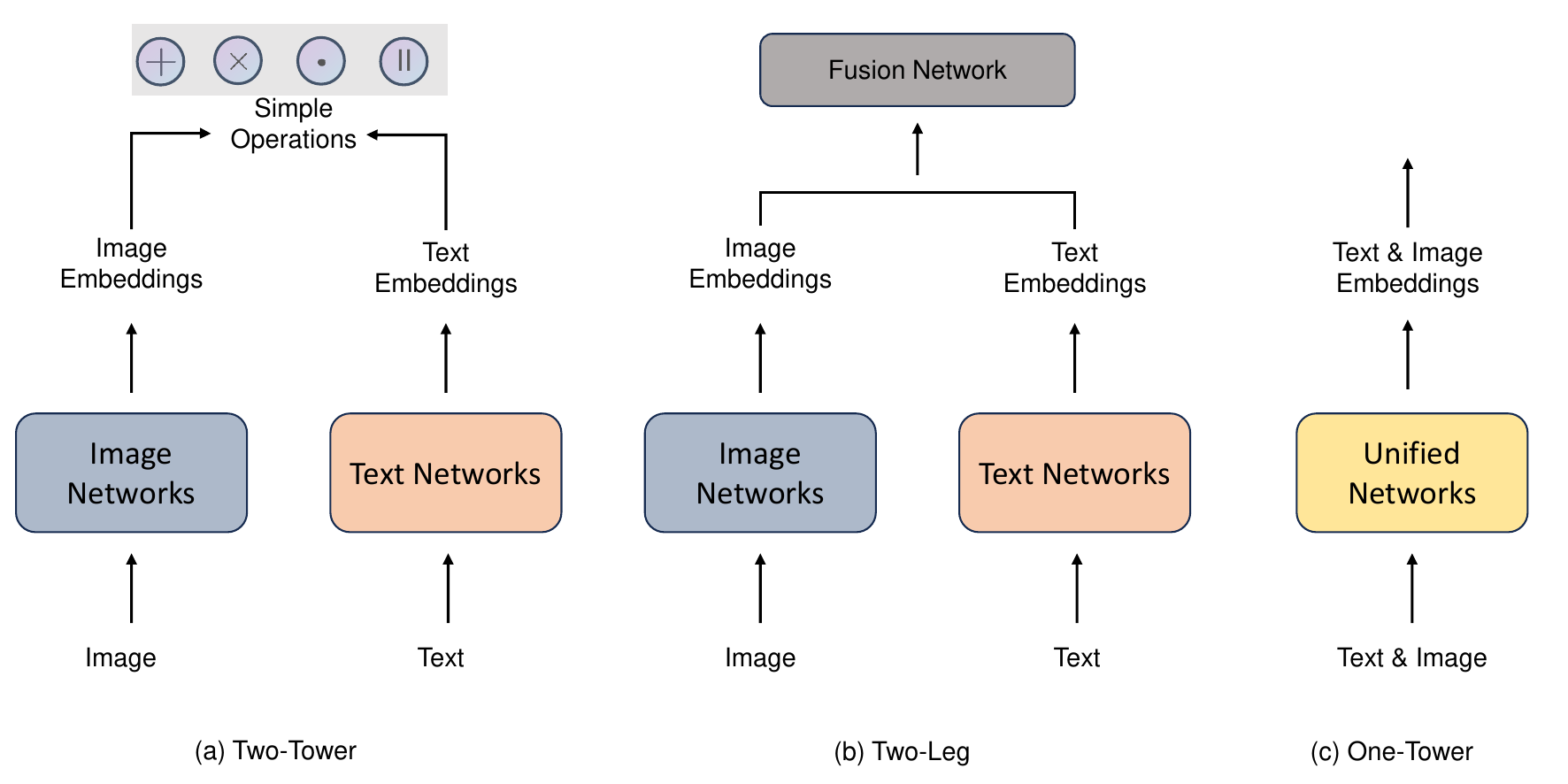}
    \caption{Overview of multimodal model architectures: (a) Two-Tower \cite{radford_clip_2021,jia2021ALIGN,liang2024multimodal,vasilakis2024instrument,xu2023bridgetower,su2023beyond,du2023touchformer,chen2024mixtower,fei2022towards,wen2024multimodal,tu2022crossmodal,yuan2021medication}: processes images and text separately, combining embeddings through simple operations (add, multiple, dot product and concatenate); (b) Two-Leg \cite{Allaire2012FusingIF,Badrinarayanan2015SegNetAD,danapal2020sensofusion,Guo2023AMF,Jaiswal2015LearningTC,Li2020HierarchicalFF,Li2018DenseFuseAF,Mai2019DivideCA,Makris2011AHF,Missaoui2010ModelLF,Rvid2019TowardsRS,Steinbaeck2018DesignOA,Uezato2020GuidedDD,Wei2021DecisionLevelDF,kim_vilt_2021}: combines separate image and text embeddings using a Fusion Network; (c) One-Tower \cite{bao_vlmo_nodate,li_blip_2022,Li2023BLIP2,chen2023instructblip,chen2023instructblip2,zhu2023minigpt4enhancingvisionlanguageunderstanding,wang2022simvlm,wang2024qwen2vlenhancingvisionlanguagemodels,bai2023qwenvl,alayrac_flamingo_2022}: utilizes a unified network to jointly embed image and text inputs.}
    \label{fig:overview}
\end{figure*}

To illustrate how modern architectures approach these challenges, Figure~\ref{fig:overview} provides an overview of three typical multimodal model structures: (a) \emph{Two‐Tower}, which processes modalities separately and combines embeddings via simple operations; (b) \emph{Two‐Leg}, which introduces a dedicated fusion network on top of separate encoders; and (c) \emph{One‐Tower}, which jointly encodes all modalities in a unified network.

While prior surveys often categorize fusion either by the stage of integration (e.g., data-level, feature-level, output-level fusion) \cite{tadas2019multimodal,barua_systematic_2023}, our work complements rather than replaces existing taxonomies. We retain the traditional classification while introducing a new organization based on the core methods they employ—ranging from statistical and kernel-based techniques to generative, contrastive, attention-based, and LLM-driven frameworks. This approach foregrounds methodological innovations and highlights how each paradigm contributes to deeper and more flexible multimodal integration. Additionally, unlike prior surveys that focus on specific modalities or models—such as vision–language models—and often treats alignment and fusion methods as a part of their survey, our work adopts a structure-centric and method-driven perspective, emphasizing general‑purpose alignment and fusion methods \cite{10445007,Zhou_2023_CVPR,li2025surveystateartlarge}. Note that while vision-text research dominates the current literature due to data availability and historical development patterns, the structural and methodological frameworks we present are designed to be transferable across modality combinations.

The organization of this survey is as follows. Section \ref{sec:prelimiaries} presents an overview of the foundational concepts in multimodal learning, including recent advances in LLMs and vision models, laying the groundwork for discussions on fusion and alignment. Section \ref{sec:why alignment and fusion} focuses on why to conduct a survey on alignment and fusion. Section \ref{sec:alignment} examines multimodal alignment and fusion approaches, presenting a dual perspective that combines structural categorizations—such as data-level, feature-level, and output-level fusion—with classifications based on the core methods and features involved in modern models. The section heavily focuses later classification from traditional strategies to recent advances, including statistical, kernel-based, graphical, generative, contrastive, attention-based and llm-based fusion frameworks, highlighting how these techniques enable deeper inter-modal integration and more flexible modeling of complex relationships. Section \ref{sec:challenge} addresses key challenges in multimodal fusion and alignment, including feature alignment, computational efficiency, data quality, and scalability. Finally, section \ref{sec:conclusion} outlines the potential directions for future research and discusses practical implications, with the aim of guiding further innovation in the field.

\section{PRELIMINARIES}
\label{sec:prelimiaries}


This section provides a brief overview of key topics and concepts to enhance the understanding of our work.

\subsection{MLLM}

Recently, both natural language processing (NLP) and computer vision (CV) have experienced rapid development, especially since the introduction of attention mechanisms and the Transformer \cite{vaswani2017attention,islam2023comprehensive,zhang2023enlighten,Tianyang2022Asurvey,zhang2024redundancy,ni2024survey,kong2025autovit,tang2024graph,ding2022looking}. Building on this framework, numerous large language models (LLMs) have emerged, such as OpenAI's GPT series \cite{radford_language_nodate,brown_language_2020,openai_gpt-4_2024} and Meta's Llama series \cite{dubey_llama_2024}. Similarly, in the vision domain, large vision models (LVMs) have been proposed, including Segment Anything \cite{kirillov2023segment}, DINO \cite{zhang2022dinodetrimproveddenoising}, and DINOv2 \cite{oquab2024dinov2learningrobustvisual}.

However, these LLMs struggle to understand visual information and handle other modalities, such as audio or sensor inputs, while LVMs have limitations in reasoning \cite{yin_survey_2023}. Given their complementary strengths, LLMs and LVMs are increasingly being combined, leading to the emergence of a new field called multimodal large language models (MLLMs). To extend the strong performance of LLMs in text processing to tasks involving other modalities, significant research efforts have been dedicated to developing large-scale multimodal models.

To extend the strong performance of LLMs in text processing to tasks involving other modalities, significant research efforts have focused on the development of large-scale multimodal models \cite{Han2024OneLLM}. Kosmos-2 \cite{peng2023kosmos2} introduces grounding capabilities by linking textual descriptions with visual contexts, allowing more accurate object detection and phrase recognition. PaLM-E \cite{driess2023palme} further integrates these capabilities into real-world applications, using sensor data for embodied tasks in robotics, such as sequential planning and visual question answering. Additionally, models like ContextDET \cite{zang2023contextdet} excel in contextual object detection, overcoming previous limitations in visual-language association by directly linking visual elements to language inputs.

Several models have adopted a hierarchical approach to managing the complexity of multimodal data. For example, SEED-Bench-2 benchmarks hierarchical MLLM capabilities, providing a structured framework to evaluate and improve model performance in both perception and cognition tasks \cite{li2023seedbench2}. Furthermore, X-LLM enhances multimodal alignment by treating each modality as a ``foreign language'', allowing a more effective alignment of audio, visual, and textual inputs with large language models \cite{chen2023xllm}.

As MLLMs continue to evolve, foundational frameworks such as UnifiedVisionGPT enable the integration of multiple vision models into a unified platform, accelerating advancements in multimodal AI \cite{kelly2023unifiedvisiongpt}. These frameworks demonstrate the potential of MLLMs to not only leverage vast multimodal datasets but also adapt to a wide range of tasks, representing a significant step toward achieving artificial general intelligence.

\subsection{Multimodal Dataset}

\begin{table*}[!ht]
\centering
\caption{Overview of different datasets' characteristics.}
\label{tab:datasets}
\resizebox{1\linewidth}{!}{%
\begin{tabular}{
    >{\arraybackslash}m{3cm} 
    |>{\arraybackslash}m{2cm} 
    |>{\arraybackslash}m{3cm} 
    |>{\arraybackslash}m{8cm} 
}
\toprule
\textbf{Dataset} & \textbf{Size} & \textbf{Modalities} & \textbf{Features} \\
\midrule
SBU Captions \cite{NIPS2011_5dd9db5e} & 1M & Image, Text & More unique words than CC-3M but fewer captions. \\
\hline
MS-COCO \cite{lin2015microsoftcococommonobjects} & 1.64M & Image, Text & Created by having crowd workers provide captions for images. \\
\hline
YFCC-100M \cite{Thomee_2016} & 100M & Image, Text & Contains 100 million image-text pairs, unclear average match degree between text and image. \\
\hline
Flickr30k \cite{plummer2016flickr30kentitiescollectingregiontophrase} & 30k & Image, Text & Created by having crowd workers provide captions for approximately 30,000 images. \\
\hline
Visual Genome \cite{krishna2016visualgenomeconnectinglanguage} & 5.4M & Image, Text & Includes structured image concepts such as region descriptions, object instances, relationships, etc. \\
\hline
RedCaps \cite{desai2021redcapswebcuratedimagetextdata} & 12.01M & Image, Text & Distributed across 350 subreddits with a long-tail distribution. Contains the distribution of visual concepts encountered by humans in everyday life without predefined object class ontologies. Higher linguistic diversity compared to other datasets like CC-3M and SBU. \\
\hline
CC-12M \cite{changpinyo2021conceptual12mpushingwebscale} & 12.4M & Image, Text & Lower linguistic diversity compared to RedCaps. \\
\hline
WIT \cite{srinivasan2021wit} & 37.6M & Image, Text & Subset of multilingual Wikipedia image-text dataset. \\
\hline
TaiSu \cite{NEURIPS2022_6a386d70} & 166M & Image, Text & TaiSu is a large-scale, high-quality Chinese cross-modal dataset containing 166 million images and 219 million Chinese captions, designed for vision-language pre-training. \\
\hline
COYO-700M \cite{kakaobrain2022coyo-700m} & 700M & Image, Text & Collection of 700 million informative image-alt text pairs from HTML documents. \\
\hline
LAION-5B \cite{schuhmann2022laion5b} & 5.85B & Image, Text & LAION-5B is a publicly available, large-scale dataset containing over 5.8 billion image-text pairs filtered by CLIP, designed for training the next generation of image-text models. \\
\hline
DATACOMP-1B \cite{gadre2023datacomp} & 1.4B & Image, Text & Collected from Common Crawl using simple filtering. Models trained on this dataset achieve higher accuracy using fewer MACs compared to previous results. \\
\hline
RS5M \cite{zhang_rs5m_2023} & 5M & Image, Text & The RS5M dataset is a large-scale remote sensing image-text paired dataset, containing 5 million remote sensing images alongside corresponding English descriptions. \\
\hline
DEAP \cite{5871728} & 2122 samples & EEG, ECG, GSR & Contains 40 one-minute music video excerpts with continuous affect ratings from 32 participants for emotion recognition. \\
\hline
PAMAP2 \cite{6246152} & 3850505 samples & IMU, Heart Rate & Multi-sensor time-series data for 18 activities with high-resolution physiological and motion data collected from 9 subjects. \\
\hline
MHEALTH \cite{10.1007/978-3-319-13105-4_14} & 120 samples & ECG, EMG, Motion & Biometric sensor streams capturing high-intensity functional movements for medical activity monitoring from 10 subjects. \\
\hline
CH-SIMS \cite{yu-etal-2020-ch} & 2,281 videos & Text, Audio, Video & Tri-modal dataset featuring multi-modal annotations across text, audio, and video modalities for sentiment analysis, with higher linguistic diversity compared to audio-only approaches. \\
\hline
MuSe-CaR \cite{stappen2021multimodalsentimentanalysiscar} & 40 hours video & Text, Audio, Video & Multimodal dataset for sentiment analysis in car reviews containing 40 hours of user-generated video material with more than 350 reviews, combining spoken language, vocal qualities, and visual cues for comprehensive sentiment understanding. \\
\bottomrule
\end{tabular}
}
\end{table*}


Different modalities offer unique characteristics. For example, images provide visual information, but are susceptible to variations in lighting and viewpoint \cite{zhao_deep_2024}. The text data are linguistically diverse and may contain ambiguities \cite{yin_survey_2023}. Audio data conveys emotional content and other non-verbal cues \cite{tadas2019multimodal}.


Multimodal datasets are foundational for training vision-language models (VLMs) by providing large-scale paired image-text data that enable model learning across various tasks, such as image captioning, text-to-image retrieval, and zero-shot classification. Key datasets include LAION-5B, WIT, and newer specialized datasets like RS5M, which target specific domains or challenges within multimodal learning. Table~\ref{tab:datasets} summarizes the commonly used datasets and their characteristics.


For example, the LAION-5B dataset contains more than 5 billion CLIP-filtered image-text pairs, enabling researchers to fine-tune models such as CLIP (Contrastive Language-Image Pretraining) and GLIDE, supporting open-domain generation and robust zero-shot classification tasks \cite{schuhmann2022laion5b}. The WIT (Wikipedia-based Image Text) dataset, with more than 37 million image-text pairs in 108 languages, is designed to support multilingual and diverse retrieval tasks, focusing on cross-lingual understanding \cite{srinivasan2021wit}. The RS5M dataset, which consists of 5 million remote sensing image-text pairs, is optimized for domain-specific learning tasks such as semantic localization and vision-language retrieval in geospatial data \cite{zhang_rs5m_2023}. Furthermore, fine-grained datasets like ViLLA are tailored to capture complex region-attribute relationships, which are critical for tasks such as object detection in medical or synthetic imagery \cite{varma_villa_2023}.

\subsection{Characteristics and Targets}

Each modality in multimodal learning presents unique challenges. For example, image data often face issues such as lighting variations, occlusions, and perspective distortions, which can affect a model's ability to recognize objects and scenes under varying conditions \cite{zhao_deep_2024_Transaction}. Text data bring complexities due to the variability of natural language, including ambiguity, slang, and polysemy, which complicate accurate interpretation and alignment with other modalities \cite{yin_survey_2023}. Similarly, audio data is susceptible to background noise, reverberation, and environmental interference, which can distort the intended signal and reduce model accuracy \cite{rahate_multimodal_2021}.

To address these challenges, specific loss functions are employed in multimodal learning to optimize both representations and alignments. These losses define how features from different modalities should be related or transformed to achieve meaningful alignment or fusion. Notable examples include:
    
- \textbf{Contrastive Loss (and Variants)}, 
    commonly used in tasks such as image-text matching, aims to pull together semantically similar pairs while pushing apart dissimilar ones in the embedding space. This objective supports better alignment and discrimination across modalities. The basic contrastive loss is defined as:
    \begin{equation}
        \mathcal{L}_{\text{contrastive}} = (1 - y)\cdot d^2 + y \cdot \max(0, m - d)^2,
    \end{equation}
    where $d$ is the distance between embeddings of a pair, $y=1$ indicates a negative pair, and $m$ is a margin hyperparameter. 

    Supervised contrastive loss extends this idea by leveraging class labels to construct positive and negative pairs within a batch. It encourages samples from the same class to cluster tightly while separating those from different classes. Its formulation is:
    \begin{equation}
        \mathcal{L}_{\text{supcon}} = -\frac{1}{N} \sum_{i=1}^{N} \log \frac{\sum_{j \in \mathcal{P}(i)} \exp(\text{sim}(z_i, z_j)/\tau)}{\sum_{k=1}^{K} \exp(\text{sim}(z_i, z_k)/\tau)},
    \end{equation}
    where $\mathcal{P}(i)$ denotes indices of positive samples for anchor $i$, $\text{sim}(\cdot,\cdot)$ is cosine similarity, $\tau$ is a temperature parameter, and $K$ includes all samples in the batch. CLIP \cite{radford_clip_2021} shares a similar objective, using contrastive-style learning with image-text pairs as positives and cross-modal negatives.

- \textbf{Sigmoid Loss}~\cite{zhai2023sigmoidlosslanguageimage} proposes a simplified and more efficient alternative to the softmax-based contrastive loss used in models like CLIP. Instead of normalizing similarities across the entire batch via softmax, which couples the loss computation to the global batch structure, the sigmoid loss treats each image-text pair independently as a binary classification problem: matching (positive) or non-matching (negative). It applies the sigmoid function to the scaled similarity score (with a learnable temperature $t$ and bias $b$) and computes a binary cross-entropy loss. Formally, for a batch of size $n$, the loss is:
    \begin{align}
        \mathcal{L}_{\text{sigmoid}} &= -\frac{1}{n} \sum_{i=1}^{n} \sum_{j=1}^{n} \log \sigma\left( z_{ij} \cdot s_{ij} \right), \\
        \text{where } s_{ij} &= t \cdot \text{sim}(z_i^{\text{img}}, z_j^{\text{text}}) + b, \text{and } \nonumber \\
        z_{ij} &= 
        \begin{cases}
            1,  & \text{if } i=j \text{ (positive pair)} \\
            -1, & \text{otherwise (negative pair)}
        \end{cases} \nonumber
    \end{align}
    This formulation significantly improves training efficiency and scalability. First, it eliminates the need for costly all-to-all batch synchronization for softmax normalization, enabling a highly memory-efficient "chunked" implementation. Second, it decouples the batch size from the loss definition, leading to superior performance at small batch sizes and allowing stable training at extremely large batch sizes (e.g., one million). Furthermore, the introduction of a learnable bias term $b$ stabilizes training by counteracting the initial extreme imbalance between positive and negative pairs, ensuring the model starts training from a more reasonable prior. These improvements make language-image pre-training more accessible and efficient, especially with limited computational resources.

- \textbf{Cross-Entropy Loss}, 
    a widely used classification loss, calculates the divergence between predicted and true probability distributions, enabling label-driven learning across modalities. It is fundamental in supervised classification tasks, and variants such as set cross-entropy offer greater flexibility for multimodal tasks by handling multiple target answers \cite{zhou_sentiment_2021,asai_set_2018}.
    
    Cross-entropy loss is particularly effective when the goal is to map multimodal inputs into a shared label space. Given predicted logits $p$ and ground truth $y$, it is defined as:
    \begin{equation}
        \mathcal{L}_{\text{CE}} = -\sum_{c=1}^{C} y_c \log(p_c),
    \end{equation}
    where $C$ is the number of classes. In multimodal settings, cross-entropy can be applied after fusing modalities or used independently per modality to encourage consistency.

- \textbf{Reconstruction Loss}, 
    used in autoencoders and multimodal fusion tasks, aims to reconstruct input data or mask noise, making models more resilient to modality-specific distortions. This type of loss is essential for multimodal tasks requiring robust feature alignment and noise resilience, such as visual-textual and audio-visual fusion \cite{parekh_representation_2020}.
    
    Reconstruction loss serves the purpose of preserving information during modality transformation or fusion. It ensures that no critical semantic content is lost during projection into a shared space. A common form is mean squared error (MSE):
    \begin{equation}
        \mathcal{L}_{\text{rec}} = \frac{1}{N} \sum_{i=1}^{N} \|x_i - \hat{x}_i\|^2,
    \end{equation}
    where $x_i$ is the original input and $\hat{x}_i$ is the reconstructed version. This loss is especially useful in unsupervised or semi-supervised multimodal architectures.

\section{Why Multimodal Alignment and Fusion}
\label{sec:why alignment and fusion}

Alignment and fusion are two fundamental concepts in multimodal learning that, while distinct, are deeply interconnected and often mutually reinforce \cite{akhmerov2019alignment,tadas2019multimodal}. Alignment involves ensuring that the different modalities are properly matched and synchronized, making the information they convey coherent and suitable for integration. Fusion, on the other hand, refers to the combination of information from different modalities to create a unified representation that captures the essence of the data in a comprehensive way \cite{tadas2019multimodal,tian2019multimodal,shankar2022fusion}. Furthermore, many recent methods find it challenging to fusion without an alignment process \cite{li_align_2021}.

\subsection{Enhancing Comprehensiveness and Robustness}

Alignment ensures that data from different sources are synchronized in terms of time, space, or context, enabling a meaningful combination. Without proper alignment, the fusion process can result in misinterpretations or loss of crucial information \cite{barua_systematic_2023}.

Once alignment is achieved, fusion utilizes the aligned data to produce a more robust and comprehensive representation \cite{li_align_2021,tao2022df}. By integrating multiple perspectives, fusion mitigates the weaknesses of individual modalities, leading to improved accuracy and reliability.

\subsection{Addressing Data Sparsity and Imbalance}



In many real-world applications, data from certain modalities may be scarce or difficult to obtain. Alignment helps to synchronize the available data, even if limited, to ensure that it can be used effectively \cite{song2024setclip,vouitsis2024dataefficient}.

The fusion then enables the transfer of knowledge between modalities, allowing the model to leverage the strengths of one modality to compensate for the weaknesses of another. This is particularly beneficial in scenarios where one modality has abundant data, while another is limited.

\subsection{Improving Model Generalization and Adaptability}



Alignment ensures that the relationships between different modalities are well understood and accurately modeled, which is crucial for the model's ability to generalize across various contexts and applications \cite{tadas2019multimodal,barua_systematic_2023}.

Fusion improves the model's adaptability by creating a unified representation that captures the nuances of the data more effectively. This unified representation can be more easily adapted to new tasks or environments, enhancing the overall flexibility of the model \cite{tadas2019multimodal,barua_systematic_2023}.

\subsection{Enabling Advanced Applications}



Alignment and fusion together enable advanced applications such as cross-modal retrieval, where information from one modality (e.g., text) is used to search for relevant information in another modality (e.g., images) \cite{Wan_2024_CVPR}. These processes are also crucial for tasks like emotion recognition \cite{7945502}, where combining visual and auditory cues provides a more accurate understanding of human emotions compared to using either modality alone.

\section{Multimodal Alignment and Fusion}
\label{sec:alignment}


Multimodal data involves the integration of various types of information, such as images, text, and audio, which can be processed by machine learning models to improve performance across numerous tasks \cite{Liang2024Foundations,tadas2019multimodal,barua_systematic_2023,10944080}. In this context, multimodal alignment and fusion are essential techniques that aim to effectively combine information from different modalities. While early approaches often processed modalities separately with only basic integration, recent methods have evolved to better capture semantic correspondences and interactions among modalities.

At a high level, these processes involve establishing meaningful relationships between heterogeneous data sources, either explicitly or implicitly, to construct representations that reflect shared semantics \cite{nassar2017multimodal,qin2023telescopic}. Explicit strategies may rely on similarity matrices to directly measure correspondences, while implicit approaches often operate in latent spaces, learning joint representations through intermediate steps such as translation or prediction \cite{tadas2019multimodal,ma2023sinkhorn}. These mechanisms are not strictly confined to one task or another but instead contribute to the broader objective of integrating diverse signals into a coherent model.

Fusion methods, in particular, play a critical role in how modalities interact within the architecture. Traditional classifications distinguish between early fusion, which combines data at the feature level \cite{Snoek2005EarlyVL}, late fusion, which merges outputs at the decision level \cite{morvant_majority_2014}, and hybrid approaches that integrate both strategies \cite{zhao_deep_2024}. However, as technology evolves, the boundaries between these categories have become increasingly blurred. Modern architectures—such as CLIP \cite{radford_clip_2021}—utilize dual encoders with relatively shallow interaction mechanisms, which are effective for retrieval-based tasks \cite{sarker_retrieval_2022,thai_medvqa_2023}, but fall short in more complex scenarios requiring nuanced understanding \cite{zhang_multimodal_intelligence_2020,tang_matr_2022}.

For tasks like visual question answering and reasoning, deeper integration is essential to capture the interdependencies between modalities, going beyond simple concatenation or independent encoding \cite{srivastava_boltzmann_2012,zhang_deep_fusion_2021}. This has led to the development of advanced fusion techniques that operate simultaneously at multiple levels of abstraction, challenging traditional stage-based categorizations. As a result, there is a growing need for a classification framework based on the core characteristics of current fusion technologies, rather than rigid temporal distinctions. Notably, attention-based mechanisms have emerged as powerful tools in this space, warranting separate treatment due to their unique contributions and rapid evolution in recent years \cite{ak_transformer_fusion_2023}. In the following parts of this section, this paper presents two classification approaches: one follows the traditional structural categorization, while the other is based on the core methods and features involved in the models.

\subsection{Structural Perspectives: A Three-level Taxonomy}
\begin{figure}[!t]
    \centering
    \includegraphics[width=0.5\textwidth]{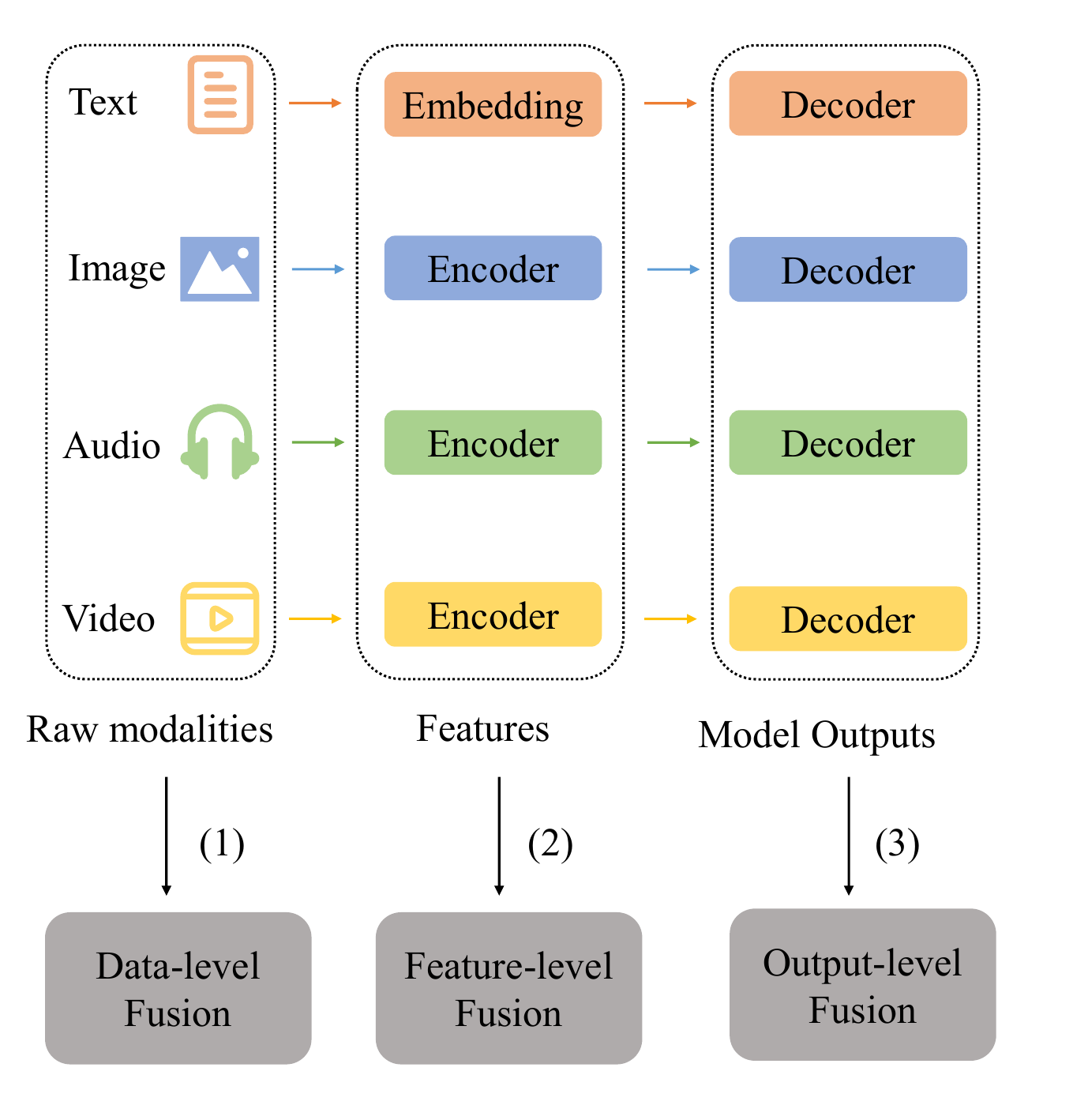}
    \caption{
    Three types of fusion with structural perspective: (1) Data-level Fusion: directly combines raw data from multiple modalities; (2) Feature-level Fusion: integrates encoded features from each modality; (3) Output-level Fusion: fuses outputs from individual modality decoders to produce a final result.}
    \label{fig:encoder-decoder}
\end{figure}

{\begin{table*}[!htbp]
\centering
\caption{Summary of models from structural perspective.}
\label{tab:multimodal_models_summary}
{
{\resizebox{1\linewidth}{!}{%
\begin{tabular}{
    >{\arraybackslash}m{2.5cm} 
    | >{\arraybackslash}m{0.6cm} 
    | >{\arraybackslash}m{1.8cm} 
    | >{\arraybackslash}m{2.1cm} 
    | >{\arraybackslash}m{7.5cm} 
}
\toprule
\textbf{Model} & \textbf{Year} & \textbf{Category} & \textbf{Modality} & \textbf{Explanation} \\
\midrule
TFN~\cite{zadeh-etal-2017-tensor} & 2017 & Output-level & Text, Audio, Image & First to use tensor fusion for high-order modality interactions at decision level. Enables end-to-end learning without intermediate fusion, capturing complex multimodal dynamics. \\
\hline
MFAS \cite{tsai2019learningfactorizedmultimodalrepresentations} & 2018 & Hybrid & Text, Audio, Image & Combines feature- and output-level fusion via factorized attention and modality gating. Dynamically weights modalities, outperforming static fusion strategies. \\
\hline
ViLBERT \cite{lu2019vilbertpretrainingtaskagnosticvisiolinguistic} & 2019 & Feature-level & Text, Image & Employs a two-stream architecture with co-attentional transformer layers for separate vision and language processing. Innovates by pretraining on masked multi-modal modeling and alignment prediction, enabling deep cross-modal interaction while preserving modality-specific processing depths. \\
\hline
UNITER \cite{chen2020uniter} & 2020 & Feature-level & Text, Image & Early unified Transformer for vision-language pretraining. Pioneered MLM, MRM, and ITM tasks, setting the standard for subsequent VLP models. \\
\hline
Perceiver \cite{jaegle2021perceivergeneralperceptioniterative} & 2021 & Data-level & Images, Point Clouds, Audio, Video & Uses asymmetric cross-attention to fuse raw inputs into a latent bottleneck. Innovates by enabling a single Transformer to handle arbitrary modalities, avoiding modality-specific architectures. \\
\hline
VX2TEXT \cite{lin2021vx2text} & 2021 & Data-level & Video, Audio, Text & Employs learnable tokenizers to convert all modalities into embeddings for unified Transformer encoding. First to unify diverse modalities end-to-end for generation, eliminating preprocessing disparities. \\
\hline
CLIP~\cite{radford_clip_2021} & 2021 & Feature-level & Text, Image & Aligns image and text via contrastive learning on large-scale data. Innovates with scalable pretraining and strong zero-shot transfer, surpassing fine-tuning-dependent models. \\
\hline
BLIP~\cite{li_blip_2022} & 2022 & Feature-level & Text, Image & Enhances CLIP with generative captioning for data refinement. Jointly optimizes understanding and generation, improving performance on both retrieval and captioning tasks. \\
\hline
FLAVA \cite{singh2022flavafoundationallanguagevision} & 2022 & Feature-level & Text, Image & Uses three separate encoders with multimodal fusion and joint unimodal/multimodal pretraining. Unique in supporting comprehensive unimodal and multimodal tasks within one model. \\
\hline
ImageBind \cite{girdhar2023imagebindembeddingspacebind} & 2023 & Feature-level & Text, Image, Audio, Depth, Thermal, IMU & Aligns six modalities in joint embedding space using image as binding medium via contrastive learning. Achieves "emergent alignment" for unseen modality pairs without direct training, enabling zero-shot cross-modal retrieval and classification. \\
\hline
ProVLA \cite{hu2023provla} & 2023 & Output-level &Text, Image & Uses two-stage Transformer with hard negative mining for progressive fusion. Improves retrieval accuracy through iterative cross-modal refinement. \\
\hline
TextBind \cite{li-etal-2024-textbind} & 2024 & Hybrid & Text, Image & Combines feature-level fusion (Q-Former mapping) and output-level fusion (LM to Stable Diffusion). Supports interleaved multimodal inputs/outputs for comprehensive instruction following with understanding and generation capabilities. \\
\bottomrule
\end{tabular}
}}
}

\end{table*}}

From structural perspectives, a multimodal model typically involves an encoder that captures essential features from the input data and compresses them into a compact form, while the decoder reconstructs the output from this compressed representation \cite{Badrinarayanan2015SegNetAD}.

In this architecture, the system is primarily composed of two major components: the encoder and the decoder. The encoder typically functions as a high-level feature extractor, transforming the input data into a latent space of significant features \cite{Badrinarayanan2015SegNetAD,Uezato2020GuidedDD}. In other words, the encoding process preserves important semantic information while reducing redundancy. Once the encoding step is complete, the decoder generates a corresponding ``reconstructed'' output based on the latent representation \cite{Badrinarayanan2015SegNetAD,Li2018DenseFuseAF}. In tasks like semantic segmentation, the decoder's output is usually a semantic label map that matches the size of the input.

In this section, we review models based on architecture. The encoder-decoder framework is an intuitive approach in which an encoder first extracts features, and then these more expressive representations are used to learn the correlations, enabling interactions between different modalities and integrating features from diverse sources. Increasingly, researchers are exploring hybrid ways to integrate features from different modalities to better reveal the relationships among them. To provide a summary, detailed information on representative models is presented in Table~\ref{tab:multimodal_models_summary}.

This encoder-decoder architecture typically takes three forms: (1) Data-level fusion, where raw data from different modalities is concatenated and fed into a shared encoder; (2) Feature-level fusion, where features are extracted separately from each modality, possibly including intermediate layers, and then combined before being input into the decoder; and (3) Output-level fusion, where outputs of individual modality-specific models are concatenated after processing. Figure~\ref{fig:encoder-decoder}  illustrates these three types of encoder-decoder fusion structures. Feature-level fusion is often the most effective, as it considers the relationships between different modalities, enabling deeper integration rather than a superficial combination.

\subsubsection{Data-level Methods}

In this method, data from each modality or processed data from each modality's unique preprocessing steps are combined at the input level \cite{danapal2020sensofusion}. After this integration, the unified input from all modalities is passed through a single encoder to extract higher-level features. Essentially, data from different modalities is merged at the input stage, and a single encoder is used to extract comprehensive features from the multimodal information.

Recent research has focused on data-level fusion to improve object detection and perception in autonomous vehicles. Studies have explored fusing camera and LiDAR data at the early stages of neural network architectures, demonstrating enhanced 3D object detection accuracy, particularly for cyclists in sparse point clouds \cite{Rvid2019TowardsRS}. A YOLO-based framework that jointly processes raw camera and LiDAR data showed a 5\% improvement in vehicle detection compared to traditional decision-level fusion \cite{danapal2020sensofusion}. Additionally, an open hardware and software platform for low-level sensor fusion, specifically leveraging raw radar data, has been developed to facilitate research in this area \cite{Steinbaeck2018DesignOA}. These studies highlight the potential of raw-data-level fusion to exploit inter-sensor synergies and improve overall system performance.

\subsubsection{Feature-level Methods}


The concept behind this fusion technique is to combine data from multiple levels of abstraction, allowing features extracted at different layers of hierarchical deep networks to be utilized, ultimately enhancing model performance. Many applications have implemented this fusion strategy \cite{tang2019fast,Scalzo2008FeatureFH,Li2020HierarchicalFF,Mai2019DivideCA,Makris2011AHF}.

Feature-level fusion has emerged as a powerful approach in various computer vision tasks. It involves combining features at different levels of abstraction to improve performance. For instance, in gender classification, a two-level hierarchy that fused local patches proved effective \cite{Scalzo2008FeatureFH}. For salient object detection, a network that hierarchically fused features from different VGG levels preserved both semantic and edge information \cite{Li2020HierarchicalFF}. In multimodal affective computing, a ``divide, conquer, and combine'' strategy explored both local and global interactions, achieving state-of-the-art performance \cite{Mai2019DivideCA}. For adaptive visual tracking, a hierarchical model fusion framework was developed to update object models hierarchically, guiding the search in parameter space and reducing computational complexity \cite{Makris2011AHF}. These approaches demonstrate the versatility of hierarchical feature fusion across various domains, showcasing its ability to capture both fine-grained and high-level information for improved performance in complex visual tasks.

\subsubsection{Output-level Methods}


Output-level fusion is a technique that improves accuracy in various applications by integrating the outputs from multiple models. For example, in landmine detection using ground penetrating radar (GPR), Missaoui et al. \cite{Missaoui2010ModelLF} demonstrated that fusing Edge Histogram Descriptors and Gabor Wavelets through a multi-stream Continuous hidden markov model (HMM) outperformed individual features and equal-weight combinations.

In multimodal object detection, Guo and Zhang \cite{Guo2023AMF} applied fusion methods such as averaging, weighting, cascading, and stacking to combine the results from models processing images, speech, and video, thereby improving performance in complex environments. For facial action unit (AU) detection, Jaiswal et al. \cite{Jaiswal2015LearningTC} found that output-level fusion using artificial neural networks (ANNs) was more effective than simple feature-level approaches.



Additionally, for physical systems involving multi-fidelity computer models, Allaire and Willcox \cite{Allaire2012FusingIF} developed a fusion methodology that uses model inadequacy information and synthetic data, resulting in better estimates compared to individual models. In quality control and predictive maintenance, a novel output-level fusion approach outperformed traditional methods, reducing prediction variance and increasing accuracy \cite{Wei2021DecisionLevelDF}. These studies demonstrate the effectiveness of output-level fusion across various domains.

\subsection{Methodological Approaches: Classification Based on Core Techniques}

\subsubsection{Statistical methods}

In the early stages, ``alignment'' often referred to the process of mapping vectors from different modalities into a shared vector space, while "fusion" typically involved a simple summation of the aligned modality vectors, followed by feeding the summed result into a neural network to obtain the final fused representation. Such alignment methods frequently relied on statistical techniques such as dynamic time warping (DTW) \cite{2007DynamicTimeWarping,kruskal1983anoverview} and canonical correlation analysis (CCA) \cite{hotelling1936cca}.

\begin{figure}[!t]
    \centering
    \includegraphics[width=0.5\textwidth]{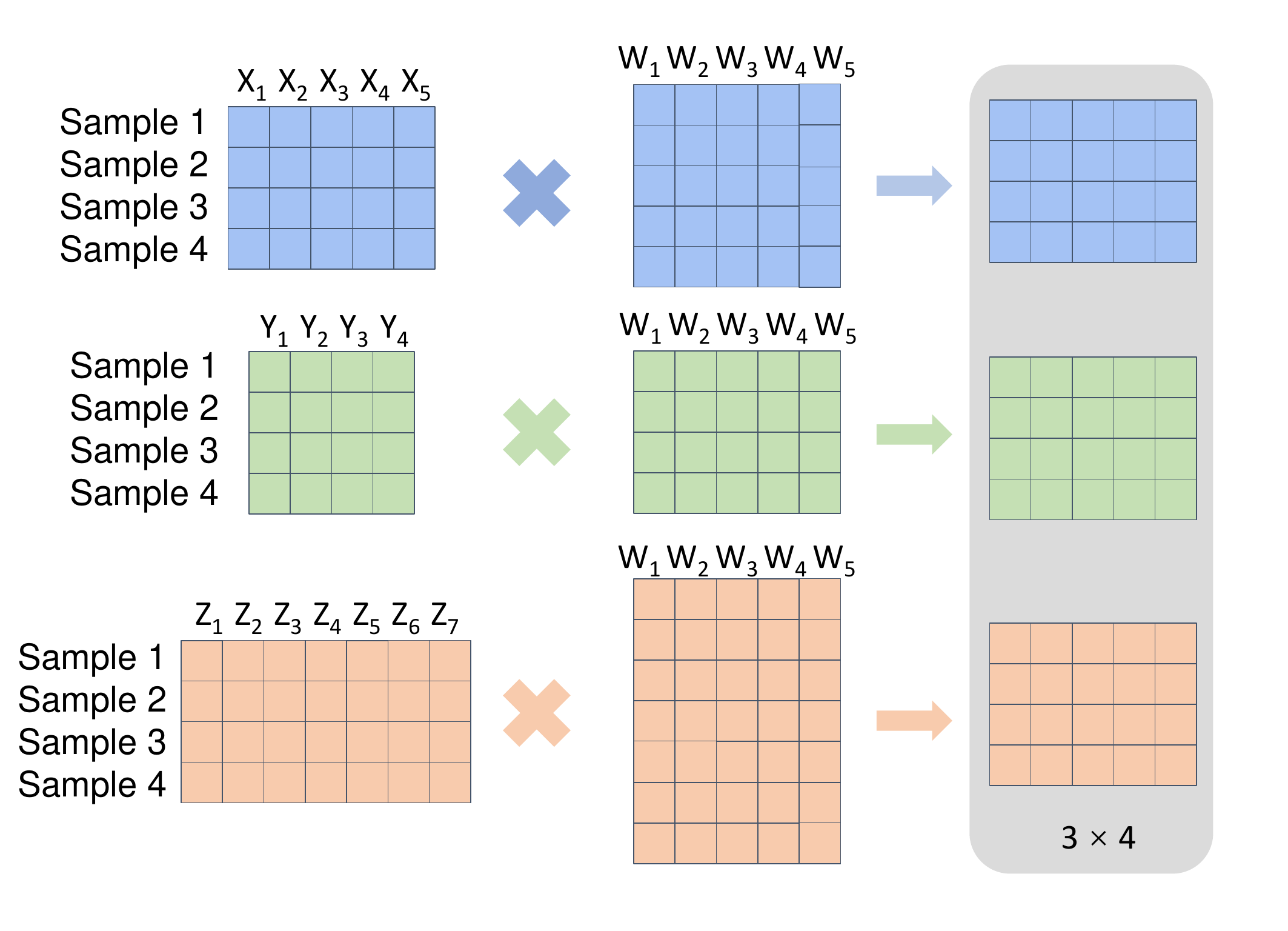}
    \caption{
    Canonical Correlation Analysis (CCA), a classic alignment method, aligns different sample matrices with varying feature dimensions using a shared weight matrix to produce a unified representation. \(X\) ,\(Y\) and \(Z\) are the data matrices from three different spaces.}
    \label{fig:CCA}
\end{figure}

DTW measures the similarity between two sequences by finding an optimal match through time warping, which involves inserting frames to align the sequences \cite{2007DynamicTimeWarping}. However, the original DTW formulation requires a predefined similarity metric, so it has been extended with CCA, introduced by Harold Hotelling in 1936 \cite{hotelling1936cca}, to project two different spaces into a common space through linear transformations. The goal of CCA is to maximize the correlation between the two spaces by optimizing the projection. CCA facilitates both alignment (through DTW) and joint learning of the mapping between modalities in an unsupervised manner, as seen in multimodal applications such as video-text and video-audio alignment. Figure~\ref{fig:CCA} visualizes the CCA method. Specifically, the objective function of CCA can be expressed as:
\begin{equation}
    \max \rho = \text{corr}(u^T X, v^T Y),
\end{equation}
where:
\begin{itemize}
    \item \(X\) and \(Y\) are the data matrices from two different spaces;
    \item \(u\) and \(v\) are the linear transformation vectors (or canonical vectors) that project \(X\) and \(Y\) into the common space;
    \item \(\rho\) is the correlation coefficient between the projections \(u^T X\) and \(v^T Y\);
    \item The goal is to find \(u\) and \(v\) that maximize the correlation \(\rho\) between the projected data.
\end{itemize}

However, CCA can only capture linear relationships between two modalities, limiting its applicability in complex scenarios involving non-linear relationships. To address this limitation, kernel canonical correlation analysis (KCCA) was introduced to handle non-linear dependencies by mapping the original data into a higher-dimensional feature space using kernel methods \cite{Bach2002kcca,Hardoon2004kcca}. Extensions such as multi-label KCCA and deep canonical correlation analysis (DCCA) further improved upon the original CCA method \cite{Akaho2001kcca,Melzer2001kcca,Bach2002kcca,Hardoon2004kcca,andrew_deep_nodate}.

Additionally, Verma and Jawahar demonstrated that multimodal retrieval could be achieved using support vector machines (SVMs) \cite{Verma2014svm}. Furthermore, methods such as linear mapping between feature modalities for image alignment have been developed to address multimodal alignment through complex spatial transformations \cite{jiang2017mapping}.

\subsubsection{Kernel-based Methods}

Kernel-based techniques have gained prominence across various domains for their ability to handle nonlinear relationships and effectively integrate heterogeneous data sources. These methods leverage the kernel trick to map data into higher-dimensional spaces, enabling improved feature representation and analysis \cite{Arany2012MultiaspectCF,Muoz2008FunctionalLO}. By selecting appropriate kernel functions, such as polynomial kernels or radial basis function kernels, these methods can achieve computational efficiency while maintaining model complexity and accuracy.

Kernel cross-modal factor analysis has been introduced as a novel approach for multimodal fusion, particularly for bimodal emotion recognition \cite{Wang2012KernelCF}. This technique identifies optimal transformations to represent coupled patterns between different feature subsets. In drug discovery, integrating multiple data sources through kernel functions within SVMs enhances drug-protein interaction predictions \cite{Wang2011KernelbasedDF}. For audio-visual voice activity detection, kernel-based fusion with optimized bandwidth selection outperforms traditional approaches in noisy environments \cite{Dov2016KernelBasedSF}. In multimedia semantic indexing, kernel-based normalized early fusion and contextual late fusion schemes demonstrate improvements over standard fusion methods \cite{Ayache2007ClassifierFF}. For drug repositioning, kernel-based data fusion effectively integrates heterogeneous information sources, outperforming rank-based fusion and providing a unique solution for identifying new therapeutic applications of existing drugs \cite{Arany2012MultiaspectCF}.

Through the use of the kernel trick, these methods achieve computational efficiency and improve prediction accuracy by better representing patterns. However, challenges exist, including difficulty in selecting the right kernel and tuning parameters, potential scalability issues with large datasets, reduced interpretability due to higher-dimensional projections, and the risk of overfitting if not properly regularized.

\subsubsection{Graphical Model-Based Methods}


The integration of graph structures allows for better modeling of complex relationships between different modalities, enabling more accurate and efficient processing of multimodal data. Such methods are commonly applied in aligning images with text or images with signals. For instance, certain models enable few-shot in-context imitation learning by aligning graph representations of objects, allowing robots to perform tasks on new objects without prior training \cite{Vosylius2023fewshot}. The GraphAlignment algorithm, based on an explicit evolutionary model, demonstrates robust performance in identifying homologous vertices and resolving paralogs, outperforming alternatives in specific scenarios \cite{Kolar2012GraphAlignment}. Figure~\ref{fig:graph-alignment} illustrates how graphs are used in alignment.

\begin{figure}[!t]
    \centering

\tdplotsetmaincoords{60}{110}
\begin{tikzpicture}[scale=2,tdplot_main_coords]

\fill[gray!10] (-1.5, -1.5, 2) -- (1.5, -1.5, 2) -- (1.5, 1.5, 2) -- (-1.5, 1.5, 2) -- cycle; 
\fill[gray!10] (-1.5, -1.5, 0) -- (1.5, -1.5, 0) -- (1.5, 1.5, 0) -- (-1.5, 1.5, 0) -- cycle; 

\foreach \i in {1,...,5} {
    \pgfmathsetmacro{\angle}{72*\i}
    \ifnum\i=4
        \node[circle,fill=blue!20,draw=black] (A\i) at ({1.5*cos(\angle)},{1.4*sin(\angle)},2) {}; 
    \else
        \ifnum\i=5
            \node[circle,fill=blue!20,draw=black] (A\i) at ({1.3*cos(\angle)},{1.5*sin(\angle)},2) {}; 
        \else
            \node[circle,fill=blue!20,draw=black] (A\i) at ({cos(\angle)},{sin(\angle)},2) {}; 
        \fi
    \fi
}

\foreach \i in {1,...,5} {
    \pgfmathsetmacro{\angle}{72*\i}
    \ifnum\i=5
        \node[circle,fill=red!20,draw=black] (B\i) at ({0.25*cos(\angle)},{0.25*sin(\angle)},0) {}; 
    \else
        \node[circle,fill=red!20,draw=black] (B\i) at ({cos(\angle)},{sin(\angle)},0) {}; 
    \fi
}

\foreach \i in {1,...,5} {
    \pgfmathtruncatemacro{\nexti}{mod(\i, 5) + 1}
    \draw (A\i) -- (A\nexti);
    \draw (B\i) -- (B\nexti);
}

\foreach \i in {1,...,3} {
    \draw[dashed] (A\i) -- (B\i); 
}

\draw (-1.5, -1.5, 2) -- (1.5, -1.5, 2) -- (1.5, 1.5, 2) -- (-1.5, 1.5, 2) -- cycle; 
\draw (-1.5, -1.5, 0) -- (1.5, -1.5, 0) -- (1.5, 1.5, 0) -- (-1.5, 1.5, 0) -- cycle; 

\node[above, fill=none, draw=none] at (0.9,0.6,2.5) {Modality 1}; 
\node[below, fill=none, draw=none] at (0,0,-0.5) {Modality 2}; 

\end{tikzpicture}
    \caption{
    In graph-based alignment, different data modalities can form graphs with distinct meanings, where the interpretation of edges and nodes may vary. For example, in \cite{Kolar2012GraphAlignment}, the interpretation of vertices and edges depends on the type of biological networks being compared.}
    \label{fig:graph-alignment}
\end{figure}

A significant challenge in these tasks is aligning implicit information across modalities, where multimodal signals do not always correspond directly to one another. Graph-based models have proven effective in addressing this challenge by representing complex relationships between modalities as graphs, where nodes represent data elements (e.g., words, objects, or frames) and edges represent relationships (e.g., semantic, spatial, or temporal) between them.

Recent studies have explored various aspects of multimodal alignment using graph structures. For instance, Tang et al. \cite{tang2021graph} introduced a graph-based multimodal sequential embedding approach to improve sign language translation. By embedding multimodal data into a unified graph structure, their model better captures complex relationships.

Another application is in sentiment analysis, where implicit multimodal alignment plays a crucial role. Yang et al. \cite{yang2023macsa} proposed a multimodal graph-based alignment model (MGAM) that jointly models explicit aspects (e.g., objects, sentiment) and implicit multimodal interactions (e.g., image-text relations).




In the domain of embodied AI, Song et al. \cite{song2023scene} explored how scene-driven knowledge graphs can be constructed to model implicit relationships in complex multimodal tasks. Their work integrates both textual and visual information into a knowledge graph, where multimodal semantics are aligned through graph-based reasoning. Aligning implicit cues, such as spatial and temporal relationships between objects in a scene, is crucial for improving decision-making and interaction in embodied AI systems.

For named entity recognition (NER), Zhang et al. \cite{zhang2021token} proposed a token-wise graph-based approach that incorporates implicit visual information from images associated with text. This method leverages spatial relations in the visual domain to improve the identification of named entities, which are often ambiguous when using isolated textual data.

In tasks such as image captioning and visual question answering (VQA), scene graphs also play a crucial role. Xiong et al. \cite{xiong2020scene} introduced a scene graph-based model for semantic alignment across modalities. By representing objects and their relationships as nodes and edges in a graph, the model improves the alignment of visual and textual modalities.



\begin{figure}[!t]
    \centering
    \resizebox{1\linewidth}{!}{
        \includegraphics{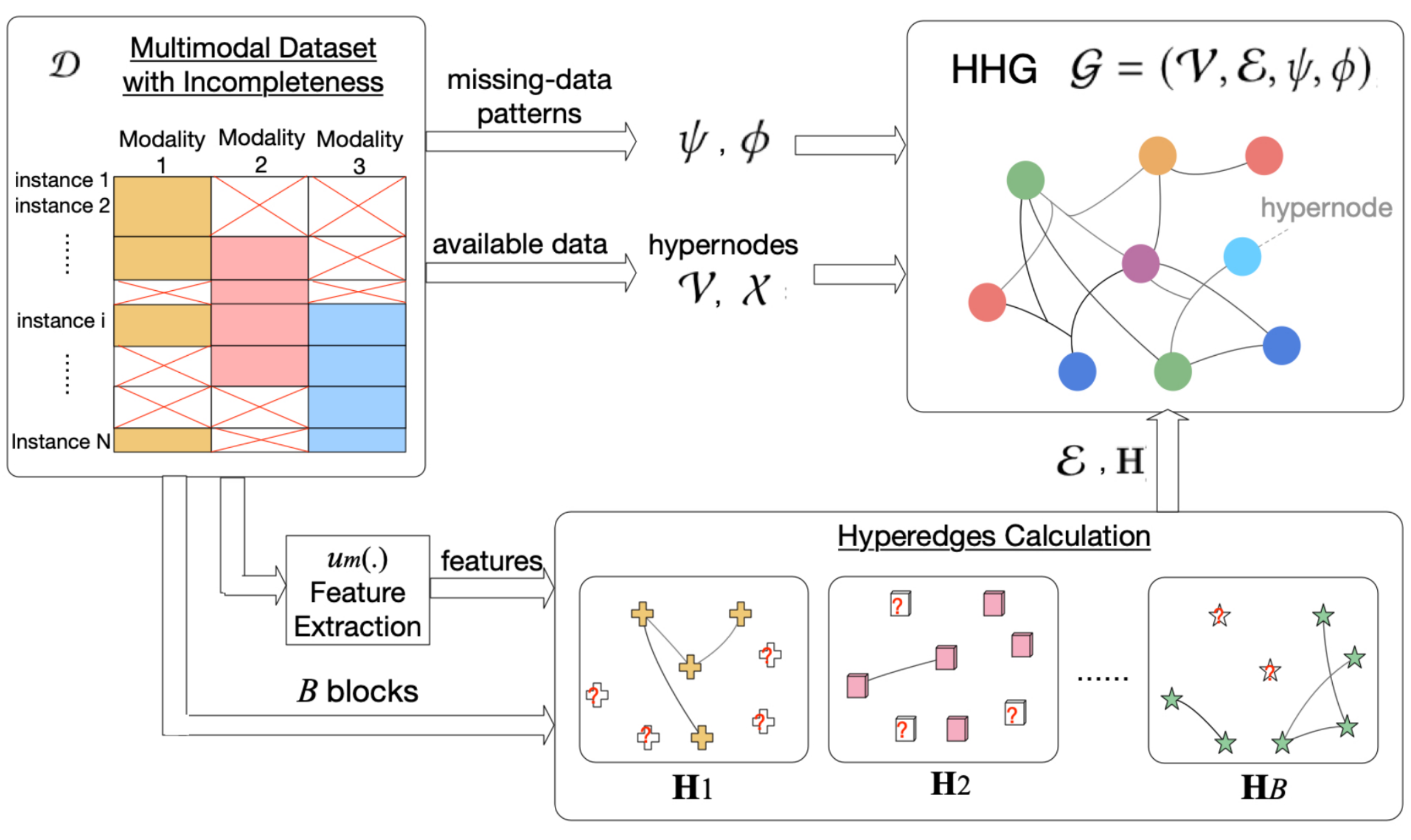}
    }
    \caption{
    Illustration from \cite{Chen2020HGMFHG}, demonstrating how graph models can effectively fuse modalities, even when some data is missing.}
    \label{fig:graph-fusion}
\end{figure}

Besides, graphical models provide a powerful approach for representing and fusing multimodal data, effectively capturing complex relationships between different modalities \cite{etin2006DistributedFI}. These models are particularly useful for handling incomplete multimodal data. For example, the heterogeneous graph-based multimodal Fusion (HGMF) method \cite{Chen2020HGMFHG} constructs a heterogeneous hypernode graph to model and fuse incomplete multimodal data. HGMF leverages hypernode graphs to accommodate diverse data combinations without requiring data imputation, enabling robust representations across various modalities \cite{Chen2020HGMFHG}. Figure~\ref{fig:graph-fusion} illustrates the construction of hypernodes in \cite{Chen2020HGMFHG}.

Graphical fusion methods are increasingly used to combine data from multiple modalities for various applications, such as Alzheimer's disease (AD) diagnosis and target tracking \cite{Blasch2014VisualizationOG,shi2022charformer}. For example, in AD diagnosis, heterogeneous graph-based models integrate neuroimaging modalities like MRI and PET, capturing complex brain network structures to improve prediction accuracy \cite{shi2021neuroimaging}. In recommendation systems, heterogeneous graphs enable the effective integration of text, image, and social media data, enhancing the quality of recommendations by capturing multimodal relationships \cite{xiong2014socialmedia}. However, traditional linear combination approaches for multimodal fusion face limitations in capturing complementary information and are often sensitive to modality weights \cite{Tong2015NonlinearGF}.

To address these issues, researchers have developed nonlinear graph fusion techniques that efficiently exploit multimodal complementarity \cite{Tong2015NonlinearGF,Tong2017MultimodalCO}. These techniques, such as early fusion operators in heterogeneous graphs, outperform linear approaches by capturing inter-modal interactions and have demonstrated improvements in one-class learning and multimodal classification tasks \cite{golo2023earlyfusion}. For instance, nonlinear fusion methods have shown enhanced classification accuracy for AD and its prodromal stage, mild cognitive impairment (MCI) \cite{Tong2017MultimodalCO}.



Recent advancements include adversarial representation learning and graph fusion networks, which aim to learn modality-invariant embedding spaces and explore multi-stage interactions between modalities \cite{Mai2019ModalityTM}. These approaches have demonstrated state-of-the-art performance in multimodal fusion tasks and provide improved visualization of fusion results \cite{Blasch2014VisualizationOG,Mai2019ModalityTM}.

In summary, graph-based methods provide a powerful framework for representing diverse data types and capturing complex, high-order interactions across modalities, making them highly effective for applications in medical diagnosis, social recommendation, and sentiment analysis. With ongoing advancements, graph-based methods hold great promise for handling incomplete, heterogeneous data and driving innovation in AI-powered multimodal applications. However, this flexibility also presents significant challenges. The sparsity and dynamic nature of graph structures complicate optimization. Unlike matrices or vectors, graphs have irregular unstructured connections, leading to high computational complexity and memory constraints. These issues persist even with advanced hardware platforms. Additionally, graph neural networks (GNNs) are particularly sensitive to hyperparameters. Choices related to network architecture, graph sampling, and loss function optimization directly impact performance, increasing the difficulty of GNN design and practical deployment.

\subsubsection{Generative Methods}
Generative methods have shown remarkable promise in learning cross-modal relationships by synthesizing and aligning high-dimensional data from different modalities. Generative adversarial networks (GANs) remain a foundational model in this domain \cite{tang2019multi,tang2021attentiongan,tang2025enhanced,tang2020local}, offering effective solutions for complex mappings between modalities. For example, DMF-GAN integrates multi-head attention and recurrent semantic fusion networks to achieve fine-grained text-to-image synthesis, significantly improving semantic alignment between modalities~\cite{yang2024dmfgan}. Similarly, GAN-based frameworks have been used for multimodal MRI synthesis, where a single generator learns unified mappings across image modalities~\cite{dai2020multimodal}.

Variational autoencoders (VAEs) also play a key role in multimodal alignment. By projecting data into shared latent spaces, VAEs enable the fusion of semantic information across modalities. This technique has proven effective in compositional tasks like image-text representation learning~\cite{wu2019multimodal}, and cross-modal quantization using VAEs has further demonstrated success in aligning text and image representations~\cite{lee2022imagetext}. 

A significant recent development in generative multimodal modeling is the adoption of diffusion models. These models offer a robust alternative to GANs and VAEs, especially in terms of stability, mode diversity, and representation fidelity~\cite{cao2025generative}. Diffusion-driven fusion techniques have enabled tasks such as face image generation from both visual prompts and text, showing strong cross-modal alignment by integrating latent representations from both GANs and diffusion models~\cite{kim2024diffusiongan}. Furthermore, semi-supervised methods like diffusion transport alignment (DTA) leverage diffusion processes for manifold alignment using minimal supervision, effectively capturing geometric similarities across modalities~\cite{duque2022diffusion}. 

More broadly, diffusion models have been successfully applied in multimodal generative frameworks like Stable Diffusion and DALL-E, enabling synthesis tasks such as image-to-audio and text-to-video generation by iteratively denoising representations conditioned on multimodal inputs~\cite{bousetouane2025vision,cao2025generative}. These approaches not only enhance the quality and coherence of generated outputs but also support flexible alignment by embedding conditional semantics at multiple stages of the generation process.

In summary, the evolution from GANs and VAEs to diffusion models marks a paradigm shift in generative multimodal fusion and alignment, offering better performance, interpretability, and multimodal coherence.

\subsubsection{Contrastive Methods}

Contrastive learning has become a cornerstone in multimodal alignment and fusion due to its ability to bring semantically related modalities closer in a shared embedding space. One of the most influential contrastive architectures is CLIP~\cite{radford_clip_2021}, which aligns text and image embeddings through large-scale pretraining on paired image-text data. CLIP and its variants have set new benchmarks in vision-language tasks, and inspired a wide range of fusion strategies.

At its core, CLIP learns aligned representations by training two encoders—one for images and one for text—so that the embeddings of matching image-text pairs are pulled closer together, while those of mismatched pairs are pushed apart. This is achieved by contrasting each sample against a batch of negatives, encouraging the model to capture semantic correspondences without requiring explicit annotations. The result is a powerful zero-shot transfer capability, where the model can generalize to unseen categories simply by encoding their textual descriptions~\cite{radford_clip_2021}.

The original CLIP model aligns global representations of images and text using a simple contrastive loss, achieving strong performance on zero-shot classification and retrieval tasks \cite{radford_clip_2021}. Recent advances have extended CLIP’s architecture to improve representation granularity and domain adaptability. For instance, HiCLIP enhances CLIP by incorporating hierarchy-aware attention in both visual and textual branches to better model fine-grained semantic relationships \cite{geng2023hiclip}. Similarly, Set-CLIP reformulates alignment as a manifold matching problem and introduces semantic density loss to improve contrastive alignment even in low-alignment settings without paired data \cite{song2024setclip}.

Other works, such as ComKD-CLIP, use contrastive distillation to transfer alignment knowledge from large CLIP models into smaller networks using image-text fusion attention mechanisms \cite{chen2024comkdclip}. SyCoCa further augments contrastive captioners by introducing bidirectional attention flows and text-guided masked image modeling to unify global and local cross-modal alignments \cite{ma2024sycoca}. Furthermore, models like Domain-Aligned CLIP explore few-shot adaptation through intra- and inter-modal contrastive learning without full finetuning \cite{gondal2023dac}.

The influence of CLIP-based models extends beyond standard vision-language domains. Applications such as image-guided editing \cite{wu2024mfeclip} and 3D representation alignment \cite{li2024gsclip} demonstrate the adaptability of contrastive alignment for diverse multimodal scenarios.

Together, these advances highlight how CLIP and its extensions form a central framework for contrastive-based multimodal alignment, driving both theoretical insight and practical performance gains.

\subsubsection{Attention-based Methods}
\label{subsec:attention-based}
\begin{figure*}[!t]
    \centering
    \resizebox{1\textwidth}{!}{
    \includegraphics[width=\textwidth]{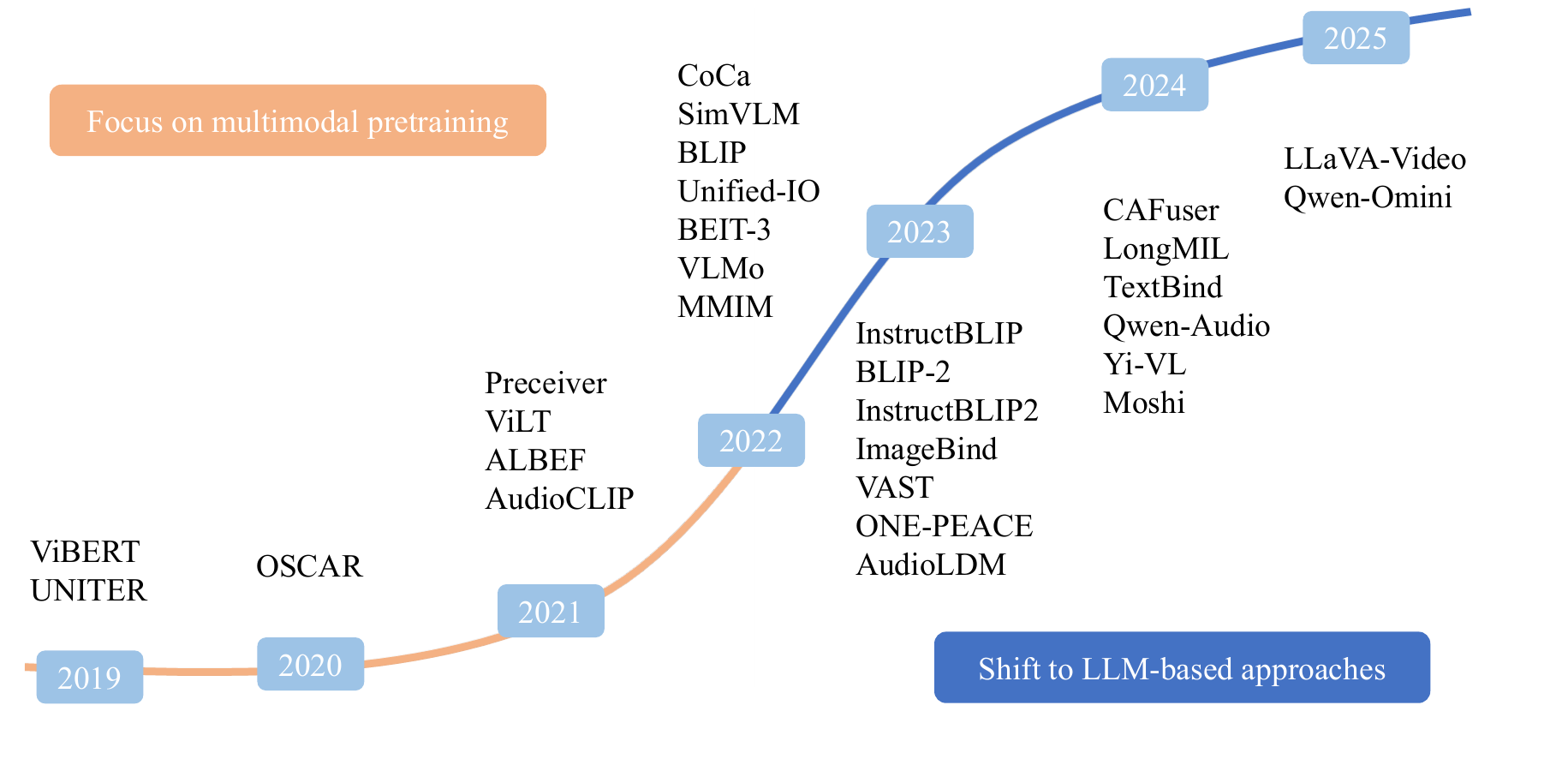}
    }
    \caption{
    Timeline of multimodal attention-based models from 2019 to 2025. From 2019 to 2022, researches heavily focued on integrate the transformer architecture for multimodal pretraining. From 2023 to 2025, the core concern has been shifted to how to reuse the knowledges of LLMs. 2019: ViLBERT~\cite{lu2019vilbertpretrainingtaskagnosticvisiolinguistic}, UNITER~\cite{chen2020uniter}; 2020: OSCAR~\cite{li2020oscar}; 2021: ViLT~\cite{kim_vilt_2021}, ALBEF~\cite{li_align_2021}, Perceiver~\cite{jaegle2021perceivergeneralperceptioniterative}, AudioCLIP~\cite{9747631}; 2022: Unified-IO~\cite{Lu2022UnifiedIO}, BEIT-3~\cite{wang_image_2022}, BLIP~\cite{li_blip_2022}, VLMo~\cite{bao_vlmo_nodate}, CoCa~\cite{yu_coca_2022}, MMIM~\cite{han-etal-2021-improving}, SimVLM~\cite{wang2022simvlm}; 2023:  ImageBind~\cite{girdhar2023imagebindembeddingspacebind}, VAST~\cite{Chen2023VAST}, ONE-PEACE~\cite{Wang2023ONEPEACE}, AudioLDM~\cite{liu2023audioldmtexttoaudiogenerationlatent}, InstructBLIP~\cite{chen2023instructblip}, BLIP-2~\cite{Li2023BLIP2}, InstructBLIP2~\cite{chen2023instructblip2}; 2024: TextBind~\cite{li-etal-2024-textbind}, CAFuser~\cite{broedermann2024condition}, LongMIL~\cite{NEURIPS2024_b7eecb72}, Qwen-Audio~\cite{chu2023qwenaudioadvancinguniversalaudio}, Yi-VL~\cite{ai2024yiopen}, Moshi~\cite{défossez2024moshispeechtextfoundationmodel}; 2025: LLaVA-Video~\cite{zhang2025llavavideovideoinstructiontuning}, Qwen-Omini~\cite{xu2025qwen25omnitechnicalreport}.}
    \label{fig:relations_attention}
\end{figure*}

Before the widespread adoption of attention mechanisms, earlier methods such as OSCAR \cite{li2020oscar}, UNITER \cite{chen2020uniter}, VILA \cite{lin2024vila}, and VinVL \cite{zhang2021vinvl} relied on object detectors to extract modality features, followed by relatively shallow fusion strategies. These pipelines lacked dynamic alignment and adaptive weighting capabilities, which are now addressed by attention-based models. Later models such as CLIP \cite{radford_clip_2021} significantly advanced image-text representation learning through contrastive pretraining.

However, the cross-modal interaction in CLIP was limited to a dot product between global embeddings, lacking fine-grained alignment or deep fusion at the token level \cite{kim_vilt_2021}. This shallow attention interaction hindered the model’s ability to fully capture complex semantic relationships across modalities, motivating the development of more integrated fusion mechanisms.

To address this limitation, methods focusing on deeper inter-modal interactions were developed, often employing Transformer encoders or other complex architectures to achieve higher-level modality integration \cite{tadas2019multimodal}. The introduction of the Vision Transformer (ViT) marked a significant shift in multimodal learning.

ViLT \cite{kim_vilt_2021} demonstrates the feasibility of performing multimodal tasks without convolutional networks or region supervision, using Transformers exclusively for feature extraction and processing. However, the simplistic structure of ViLT led to performance issues, particularly when compared to methods that emphasized deeper inter-modal interactions and fusion \cite{li_align_2021,tadas2019multimodal,yang2022continual}. ViLT lagged behind these methods in many tasks, possibly due to dataset bias or the inherent need for stronger visual capabilities \cite{li_align_2021}. Generally, visual models need to be larger than text models to achieve better results, and the performance degradation was not primarily caused by the lightweight visual embedding strategy.

Subsequent works, such as ALBEF \cite{li_align_2021}, introduced more sophisticated model designs. ALBEF emphasized aligning image and text representations before their fusion using a contrastive loss. By employing momentum distillation, it generated pseudo-labels to mitigate challenges posed by noisy datasets. Following this, BLIP \cite{li_blip_2022} adopted a bootstrapping mechanism, using initially generated captions from the model to filter out dataset noise, thereby improving the quality of subsequent training.

Attention-based mechanisms gained prominence with the introduction of the Transformer architecture \cite{vaswani2017attention} and following works \cite{devlin2019bertpretraining,zuo2023svp,grail-etal-2021-globalizing,zhang2023neuralattentionenhancingqkv}, where the attention function takes queries (Q), keys (K), and values (V) and computes the relevance of each key to a given query. The scaled dot-product attention is given by:
\begin{equation}
    \text{Attention}(Q, K, V) = \text{softmax} \left( \frac{QK^\top}{\sqrt{d_k}} \right) V.
\end{equation}
This operation dynamically weighs each input feature according to its contextual importance, enabling the model to capture long-range dependencies across sequences or modalities.

The use of attention mechanisms enables decoders to focus selectively on specific subcomponents of the source input. This contrasts with traditional encoder-decoder models that treat all source features as a single representation. Attention modules guide decoders to highlight task-relevant regions—such as specific image patches, words in a sentence, or audio frames—during output generation. For instance, in image captioning, attention allows decoders to attend to relevant visual regions when generating each word, instead of encoding the entire image as a static vector \cite{karpathy2015deep}.

Considering the inherent connection between attention score and similarity score, attention mechanisms are widely applied for alignment, i.e., learning correspondences between semantically similar elements across modalities. For example, the Att-Sinkhorn method uses the Sinkhorn distance in conjunction with attention to model cross-modal optimal transport, aligning features from different distributions \cite{ma2023sinkhorn}. The AbFTNet model emphasizes "alignment-before-fusion" by first synchronizing modality-specific features via a Transformer-based mechanism and then integrating them through a cross-modal aggregation module \cite{ning2024abftnet}. In the domain of knowledge representation, DSEA applies a dynamic self-attention network to evaluate the importance of structural and attribute information, improving entity alignment across multimodal knowledge graphs \cite{qian2023leveraging}.

Attention-based fusion integrates multimodal information by learning how much to attend to each modality. This selective integration helps manage noise, modality imbalance, and complementary cues. For instance, BMAN applies multi-head attention to align and fuse audio-text features with learnable weights, improving sentiment prediction across unaligned datasets \cite{zhang2023bimodal}. ProVLA employs a cross-attention fusion encoder that progressively aligns vision and language representations for compositional image retrieval \cite{hu2023provla}.

Attention-based fusion is particularly effective in multimodal tasks because it supports flexible integration and handles modality-specific uncertainties \cite{Tang2023BAFN,Wang2023Mutually}. However, this flexibility comes at the cost of increased computational complexity and often demands large-scale annotated datasets.

Recent advances such as ProVLA \cite{hu2023provla} and AbFTNet \cite{ning2024abftnet} take Transformer-based fusion further by introducing progressive alignment and structured attention mechanisms prior to fusion. ProVLA employs a two-stage alignment-fusion paradigm, leveraging cross-attention for robust semantic integration and using momentum-based hard negative mining to enhance alignment robustness. Similarly, AbFTNet introduces a CAP (Cross-modal Aggregation Promoting) module, aligning unimodal features through self-attention before cross-modal integration, thus addressing modality-specific information disparities.

Fusion mechanisms like TokenFusion \cite{wang2022tokenfusion} also explore token-level replacement and residual alignment to balance modality contributions and avoid attention dilution. These designs allow Transformers to retain unimodal strengths while gaining inter-modal awareness through informed token substitution and dynamic fusion.

CoCa \cite{yu_coca_2022} combined contrastive loss with captioning loss, achieving remarkable performance. In particular, CoCa excelled not only in multimodal tasks, but also performed well on single-modal tasks such as ImageNet classification. BEIT-3 \cite{wang_image_2022} further advanced multimodal learning with the implementation of Multiway Transformers, enabling the simultaneous processing of images, text, and image-text pairs. By applying masked data modeling to these inputs, BEIT-3 achieved state-of-the-art performance across various visual and vision-language tasks.

Figure~\ref{fig:relations_attention} illustrates the relationships among major works related to attention mechanisms and transformers and Table~\ref{tab:multimodal_summary} provides a summary of representative models.


\begin{table*}[!htbp]
\centering
\caption{Summary of attention-based multimodal models.}
\label{tab:multimodal_summary}
    \resizebox{\linewidth}{!}{%
\begin{tabular}{
    >{\arraybackslash}m{2.5cm} 
    |>{\arraybackslash}m{0.6cm} 
    |>{\arraybackslash}m{3cm} 
    |>{\arraybackslash}m{6.5cm} 
}
\toprule
\textbf{Model} & \textbf{Year} & \textbf{Modality} & \textbf{Training Modules} \\
\midrule
ViLBERT \cite{lu2019vilbertpretrainingtaskagnosticvisiolinguistic} & 2019 & Vision, Text & Dual-stream co-attentional transformer, pretrained on Conceptual Captions, for task-agnostic visiolinguistic representation in VQA, VCR, referring expressions, and image retrieval. \\
\hline
UNITER \cite{chen2020uniter} & 2019 & Vision, Text & MLM, MRM, ITM, and WRA via Optimal Transport for fine-grained alignment.\\
\hline
Oscar \cite{li2020oscar} & 2020 & Vision, Text & Object-Semantics Aligned Pre-training using detected object tags as anchor points. \\
\hline
ViLT \cite{kim_vilt_2021} & 2021 & Vision, Text & Masked Language Modeling; Image-Text Matching; Word-Patch Alignment. \\
\hline
ALBEF \cite{li_align_2021} & 2021 & Vision, Text & Image-Text Contrastive Loss; Image-Text Matching Loss; Masked-Language-Modeling Loss. \\
\hline
Perceiver \cite{jaegle2021perceivergeneralperceptioniterative} & 2021 & Arbitrary Modalities & General-purpose architecture handling sequences of varying modalities. \\
\hline
AudioCLIP \cite{9747631} & 2021 & Audio, Vision, Text & Contrastive learning extended from CLIP to audio domain. \\
\hline
Unified-IO \cite{Lu2022UnifiedIO} & 2022 & Vision, Text & Object Segmentation; Visual Question Answering; Depth Estimation; Object Localization. \\
\hline
BEIT-3 \cite{wang_image_2022} & 2022 & Vision, Text & Masked ''language'' modeling for images, texts, and image-text pairs. \\
\hline
BLIP \cite{li_blip_2022} & 2022 & Vision, Text & Image-Text Contrastive Loss; Image-Text Matching Loss; Language Modeling Loss. \\
\hline
VLMo \cite{bao_vlmo_nodate} & 2022 & Vision, Text & Image-Text Contrastive Learning; Masked Language Modeling; Image-Text Matching. \\
\hline
CoCa \cite{yu_coca_2022} & 2022 & Vision, Text & Captioning Loss; Contrastive Loss. \\
\hline
MMIM \cite{han-etal-2021-improving} & 2022 & Audio, Text, Non-Verbal Context & Multi-task learning with unobserved multimodal context for sentiment analysis. \\
\hline
ImageBind \cite{girdhar2023imagebindembeddingspacebind} & 2023 & Vision, Text, Audio, Depth, Thermal, IMU & Zero-shot alignment across 6 modalities via image-centric binding. \\
\hline
VAST \cite{Chen2023VAST} & 2023 & Vision, Text & OM-VCC; OM-VCM; OM-VCG \\
\hline
ONE-PEACE \cite{Wang2023ONEPEACE} & 2023 & Vision, Audio, Text & Masked Contrastive Learning; supports multi-modal fusion through self-attention layers. \\
\hline
Coupled Mamba \cite{li2024coupledmambaenhancedmultimodal} & 2024 & Vision, Text, Audio & Enhanced multimodal fusion with State Space Models (SSMs) for non-LLM architectures. \\
\hline
CAFuser \cite{broedermann2024condition} & 2024 & Vision, Text & Image-Text Contrastive Loss \\
\hline
LongMIL \cite{NEURIPS2024_b7eecb72} & 2024 & Medical Images, Text & Local-Global Hybrid Transformer architecture for long-context multiple instance learning; Self-supervised contrastive learning for medical image-text alignment. \\
\bottomrule
\end{tabular}
}
\end{table*}

\subsubsection{LLM-based Methods}

\begin{figure*}[ht]
    \centering
    \includegraphics[width=\textwidth]{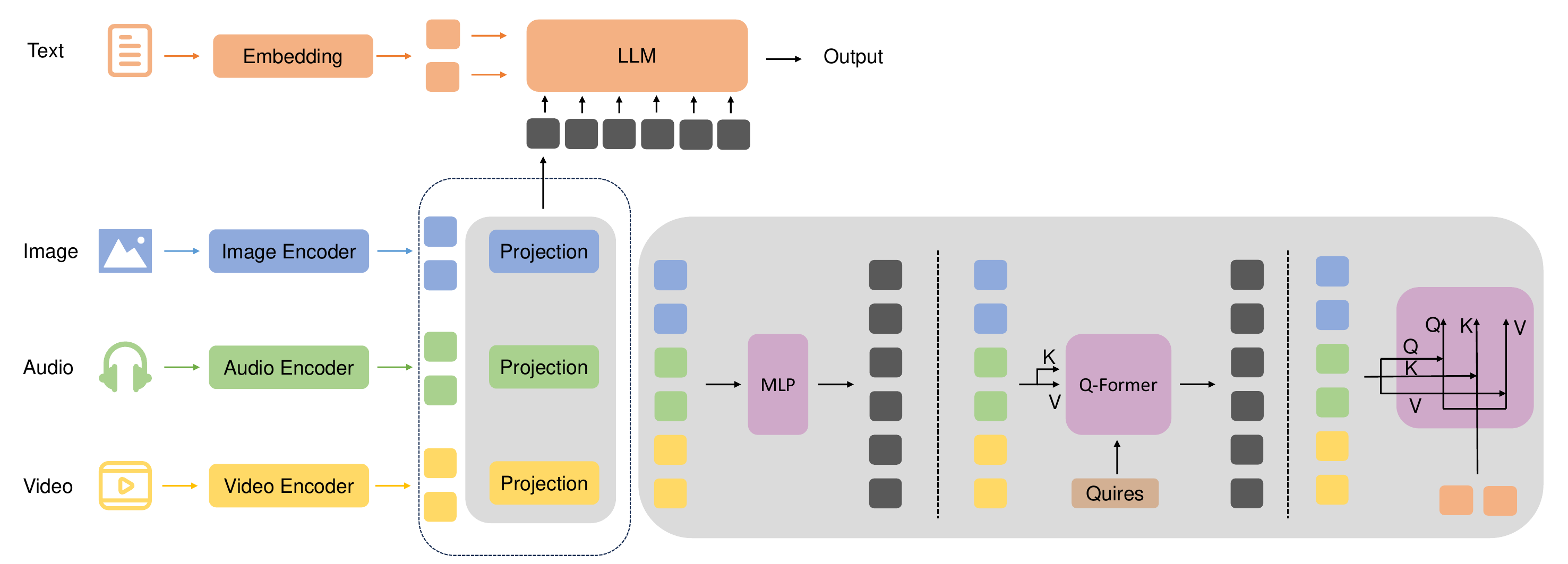}
    \caption{
    This pipeline demonstrates multimodal fusion using a large language model (LLM). Text inputs are embedded and processed by the LLM, while image, audio, and video inputs are encoded, projected into a shared embedding space, and passed through modules such as an MLP and Q-Former. The Q-Former uses attention mechanisms (queries, keys, and values) to align multimodal features before generating a final output through the LLM.
    }
    \label{fig:deep_fusion}
\end{figure*}

\begin{figure}[t]
    \centering
    \includegraphics[width=0.5\textwidth]{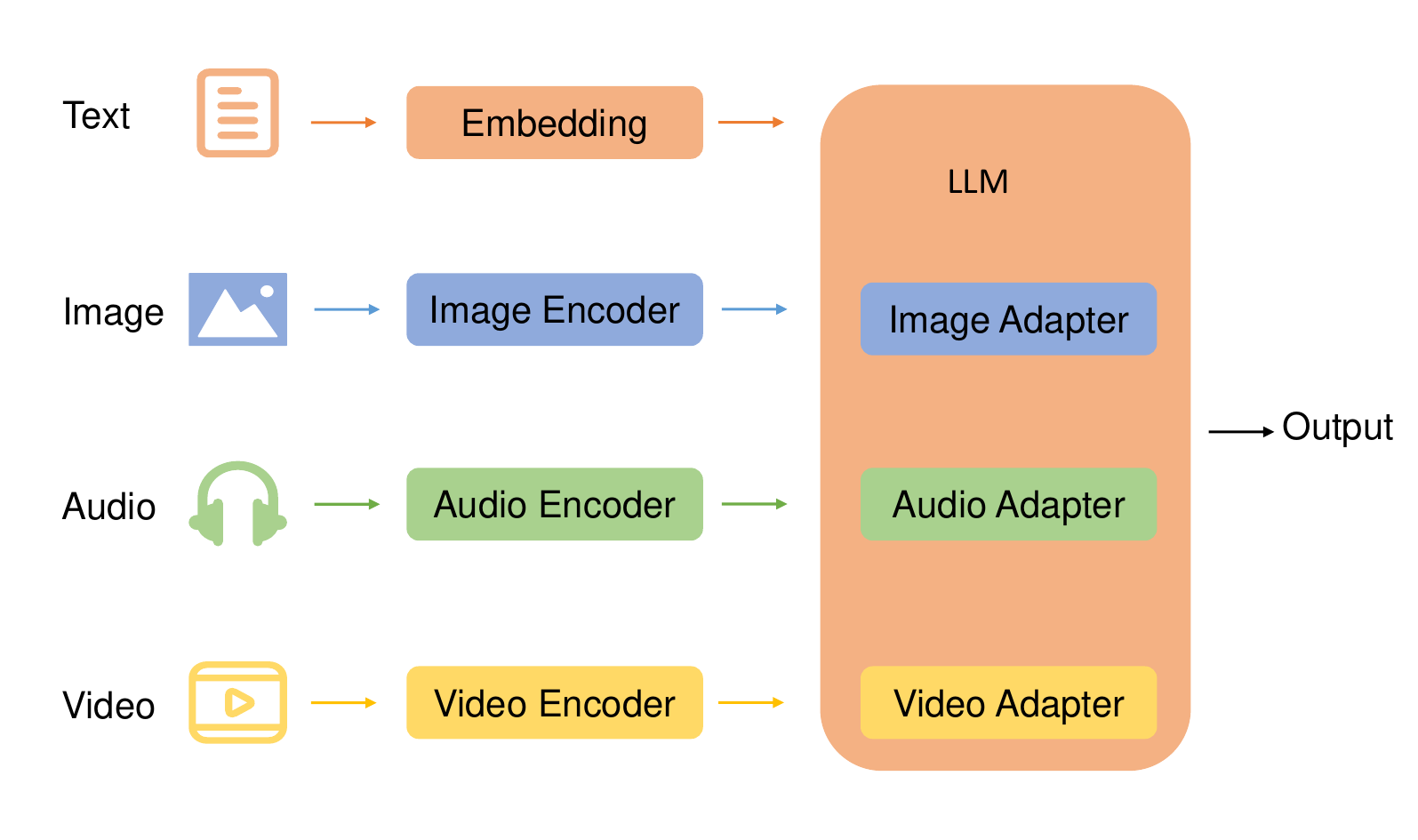}
    \caption{
    Adapters embedded in LLM. Each modality (text, image, audio, and video) is processed by its respective encoder and a modality-specific adapter. These adapters then feed the encoded features into an LLM, which generates the final output.
    }
    \label{fig:adapter}
\end{figure}

As mentioned in Section~\ref{subsec:attention-based} and Figure~\ref{fig:relations_attention}, the trend has shifted to LLM-based models from 2022. Figure~\ref{fig:deep_fusion} illustrates a common scenario of LLM-based method. After the encoder extracts features from each modality, a connector maps these features into the text space, where they are processed together by the LLM. Previously, this connector was often a simple MLP, but it can now be a more complex attention mechanism. Recently, researchers have proposed various architectures and techniques aimed at enhancing cross-modal capabilities. They embed adapters into frozen LLMs to facilitate interactions between modalities~\cite{NEURIPS2023_5e84e441}. Figure~\ref{fig:adapter}  shows the typical structure of this approach. The key difference from previous methods is that adapters are embedded directly into the LLMs, allowing for end-to-end training with alignment included.
For example, the Qwen-VL series models \cite{bai2023qwenvl} advanced cross-modal learning through the design of visual receptors, input-output interfaces, and a multi-stage training pipeline, achieving notable performance in image and text understanding, localization, and text reading. In video understanding, the ViLA network \cite{Wang2024ViLA} introduced a learnable text-guided Frame-Prompter and a cross-modal distillation module (QFormer-Distiller) optimized for key frame selection, improving both accuracy and efficiency in video-language alignment. Additionally, CogVLM \cite{wang_cogvlm_2024} incorporated visual expertise into pretrained language models using Transformers. In emotion recognition tasks, COLD Fusion added an uncertainty-aware component for multimodal emotion recognition \cite{tellamekala_cold_2024}.

Various pre-training strategies have been developed to facilitate multimodal fusion. For example, BLIP-2 \cite{Li2023BLIP2} introduced a bootstrapping approach that used frozen image encoders and large language models for vision-language pre-training, reducing the number of parameters while enhancing zero-shot learning performance. Similarly, the VAST model \cite{Chen2023VAST} explored a comprehensive multimodal setup involving vision, audio, subtitles, and text, constructing a large-scale dataset and training a foundational model capable of perceiving and processing all these modalities. Furthermore, the ONE-PEACE model \cite{Wang2023ONEPEACE} employed a modular adapter design and shared self-attention layers to provide a flexible and scalable architecture that could be extended to more modalities. The research by Zhang et al. \cite{Zhang2022TransformerBasedEA} used Transformers for end-to-end anatomical and functional image fusion, leveraging self-attention to incorporate global contextual information.

Despite these advances, the field still faces several challenges. One of the main challenges is data bias, where inherent biases in training datasets limit model performance. Another concern is maintaining consistency across modalities to ensure coherent information integration without loss or inconsistency. Additionally, as models grow in scale, there is an increasing demand for computational resources, necessitating more efficient algorithms and hardware support. Table~\ref{tab:llm_based_summary} summarizes some state-of-the-art (SOTA) or popular LLM-based models.

In conclusion, multimodal fusion remains a dynamic and evolving area of research, driven by advances in attention-based mechanisms and model architectures. Although significant progress has been made in developing models that effectively integrate information from multiple modalities, ongoing challenges such as data bias, modality consistency, and computational demands persist. Continued exploration of new theoretical frameworks and technical solutions is necessary to achieve more intelligent and adaptable multimodal systems, advancing artificial intelligence technologies, and providing powerful tools for practical applications.

\begin{table*}[!htbp]
\centering
\caption{Summary of LLM-based multimodal models with diverse modalities.}
\label{tab:llm_based_summary}
\resizebox{\linewidth}{!}{%
\begin{tabular}{
>{\arraybackslash}m{2.5cm} 
|>{\arraybackslash}m{0.6cm} 
|>{\arraybackslash}m{2.4cm} 
|>{\arraybackslash}m{2cm} 
|>{\arraybackslash}m{7cm} 
}
\toprule
\textbf{Model} & \textbf{Year} & \textbf{Modality} & \textbf{Used LLM} & \textbf{Training Modules} \\
\midrule
MiniGPT-4 \cite{zhu2023minigpt4enhancingvisionlanguageunderstanding} & 2023 & Text, Image & Vicuna & Two-stage training: Stage 1: Freeze visual feature extractor, train projection layer to align visual features with Vicuna; Stage 2: Instruction finetuning on dialogue data \\ \hline
Qwen-VL \cite{bai2023qwenvl} & 2023 & Text, Image & Qwen-7B & Stage 1: Image caption generation; Stage 2: Multitask pretraining; Stage 3: Supervised finetuning \\ \hline
BLIP-2 \cite{Li2023BLIP2} & 2023 & Text, Image & OPT, FlanT5 & Stage 1: Vision-Language Representation Learning; Stage 2: Vision-to-Language Generation Learning \\ \hline
LLaVA \cite{liu2023visualinstructiontuning} & 2023 & Text, Image & GPT-3, GPT-3.5, LLaMA & Visual Instruction Tuning \\ \hline
LaVIN \cite{NEURIPS2023_5e84e441} & 2023 & Text, Image & LLaMA & Fine-tuning with MoE adapter \\ \hline
MiniGPT-v2 \cite{chen2023minigptv2largelanguagemodel} & 2023 & Text, Image & Vicuna (7B/13B) & Multitask learning \\ \hline
InstructBLIP \cite{chen2023instructblip} & 2023 & Text, Image & Vicuna (7B/13B) & Visual Instruction Tuning \\ \hline
InternLM-XComposer \cite{zhang2023internlmx} & 2023 & Text, Image & InternLM-Chat-7B & Pre-training, Multi-task Training, Instruction Fine-tuning \\ \hline
Macaw-LLM \cite{lyu2023macawllmmultimodallanguagemodeling} & 2023 & Text, Image, Audio, 3D & LLaMA & Multimodal language modeling with unified representation \\ \hline
3D-MMLM \cite{hong20233dllminjecting3dworld} & 2024 & Text, Image, 3D & LLaMA-3 & 3D understanding, point cloud processing, cross-modal alignment \\ \hline
Qwen2-VL \cite{wang2024qwen2vlenhancingvisionlanguagemodels} & 2024 & Text, Image & Qwen-2 & Visual Instruction Tuning \\ \hline
Moshi \cite{défossez2024moshispeechtextfoundationmodel} & 2024 & Text, Audio & LLaMA-2 & Text-to-speech, speech-to-text, audio understanding \\ \hline
MM-LLMs \cite{zhang-etal-2024-mm} & 2024 & Text, Image, Video, Audio, 3D & Various & Handling different modalities where X can be image, video, audio, 3D, etc. \\ \hline
AudioPaLM \cite{rubenstein2023audiopalmlargelanguagemodel} & 2023 & Text, Audio & PaLM-2 & Speech recognition, speech synthesis, multilingual audio understanding \\ \hline
Qwen-Audio \cite{chu2023qwenaudioadvancinguniversalaudio} & 2024 & Text, Audio & Qwen-1.5 & Audio instruction tuning, audio-text alignment \\ \hline
Yi-VL \cite{ai2024yiopen} & 2024 & Text, Image & Yi-Chat & Three-stage training: 1. Train ViT and projection module; 2. Increase image resolution and retrain; 3. Train entire model \\ \hline
InternLM-XComposer-2.5 \cite{zhang2024internlmx2.5} & 2024 & Text, Image & InternLM2-7B & Pre-training, Multi-task Training, Instruction Fine-tuning \\ \hline
CogVLM \cite{wang_cogvlm_2024} & 2024 & Text, Image & LLaMA-2 & Pre-training + Supervised Fine-tuning on vision-language tasks \\ \hline
ViLA \cite{Wang2024ViLA} & 2024 & Text, Image & Supports Frozen and Finetuned (LoRA) usage of LLM & Distillation loss; Visual Question Answering loss \\ \hline
LLaVA-Video \cite{zhang2025llavavideovideoinstructiontuning} & 2025 & Text, Image, Video & LLaMA-2 & Video instruction tuning, video question answering \\ \hline
Qwen2.5-Omni \cite{xu2025qwen25omnitechnicalreport} & 2025 & Text, Image, Audio, Video, 3D & Qwen-2.5 & Video and audio branches with cross-attention for multimodal understanding \\ 
\bottomrule
\end{tabular}
}
\end{table*}

\section{Challenges in Multimodal Alignment and Fusion}
\label{sec:challenge}

\subsection{Multimodal Misalignment and Modality Gap}

Multimodal misalignment and modality gap are two critical challenges in multimodal representation learning that significantly affect model performance. Misalignment refers to the mismatch between different modalities, such as images and their corresponding textual descriptions, which can arise due to noisy or incorrect annotations~\cite{tong2024eyeswideshutexploring}. For instance, Ma et al. \cite{ma_learning_2024} identified modality misalignment as a significant barrier to transferring knowledge across different modalities, emphasizing that pre-trained models frequently struggle with knowledge transfer when there is a substantial semantic gap between modalities.

The modality gap, on the other hand, describes the disparity in embedding distributions of different modalities within a shared space, leading to suboptimal cross-modal interactions~\cite{Liang2022MindTG}. The modality gap in multimodal contrastive representation learning arises due to a combination of factors. First, deep neural networks inherently create a ``cone effect'', where embeddings from a single modality are restricted to a narrow region of the embedding space. This geometric bias is amplified by nonlinear activation functions and network depth. Second, different random initializations for separate encoders in multimodal models result in distinct cones in the shared embedding space, causing a separation between modalities even at initialization. Finally, during training, contrastive learning preserves this gap. The contrastive loss, especially at low temperatures, maintains a repulsive structure in the optimization landscape, preventing the gap from closing. Together, these factors explain the persistent separation between modalities observed in multimodal models like CLIP~\cite{Liang2022MindTG}.

Many approaches seek to mitigate the modality gap through various architectural and training innovations. For instance, noise-injected embeddings, such as those used in CapDec, improve generalization by perturbing CLIP embeddings to reduce overfitting to specific modality characteristics and enhance alignment in low-resource settings~\cite{nukrai2022capdec}. Meanwhile, VT-CLIP improves modality alignment by incorporating visual-guided text generation to highlight image regions that correspond to key linguistic cues~\cite{zhang_vtclip_2021}. Finite discrete tokens (FDT) aim to resolve the granularity gap between visual patches and textual tokens by embedding both into a unified semantic space~\cite{chen2023contrastive}. Modality knowledge alignment (MoNA) introduces a meta-learning paradigm to learn target data transformations that reduce modality knowledge discrepancies prior to transfer~\cite{ma_learning_2024}. Recently, Tong et al.~\cite{tong2024eyeswideshutexploring} investigate the gap between the visual embedding space of CLIP and vision-only self-supervised learning, proposing a mixture of features (MoF) approach to enhance visual grounding capabilities.

To advance this field, future research should focus on developing more sophisticated embedding space modeling techniques, dynamically adjusting modality gaps during training, and improving data quality through better annotation practices. Additionally, enhancing the compositional understanding of VLMs by incorporating syntactic structure and word order sensitivity—such as generating targeted negative captions through linguistic element swapping—could further improve model capabilities~\cite{yuksekgonul2023visionlanguagemodelsbehavelike}. These directions aim to bridge existing gaps in multimodal learning and push models toward a more human-like comprehension of multimodal input.

\subsection{Computational Efficiency Challenge}




Early multimodal models faced significant computational demands due to their reliance on object detectors, particularly during inference. The introduction of vision transformers (ViTs) enabled patch-based visual representations instead of bounding boxes, reducing computational complexity. However, simply tokenizing textual and visual features remains insufficient for effective multimodal processing. Recent work has proposed several efficient fusion mechanisms to mitigate this challenge. 

TokenFusion, for instance, dynamically replaces less informative tokens with fused inter-modal features to reduce token redundancy in transformer architectures~\cite{wang2022tokenfusion}. Attention bottlenecks allow selective modality interaction through shared tokens, enabling low-rank communication between modalities with minimal overhead~\cite{nagrani2021bottlenecks}. Prompt-based multimodal fusion (PMF) introduces deep-layer prompts that interact across pretrained unimodal Transformers, achieving comparable performance to full fine-tuning~\cite{li2023pmf}. Other notable methods include sparse fusion transformers (SFT), which sparsify unimodal tokens prior to fusion~\cite{ding2021sparsefusion}, and dynamic multimodal fusion (DynMM), which adaptively determines forward computation paths using data-dependent gating~\cite{xue2022dynamicfusion}. Low-rank tensor fusion approaches have also been explored, leveraging compact representations to avoid exponential parameter growth~\cite{liu2018lowrank}. Recently, GeminiFusion proposed a pixel-wise fusion method with linear complexity by combining intra- and inter-modal attention, controlled by layer-adaptive noise modulation~\cite{jia2024geminifusion}.

Despite these advances, fusion remains the computational bottleneck in large-scale multimodal models. Future research should focus on developing adaptive, scalable, and resource-efficient fusion architectures to meet the growing demand of real-world multimodal tasks.

\subsection{Data Quality and Availability}

One of the primary obstacles is the scarcity of large-scale, high-quality multimodal datasets, which are essential for training robust multimodal language models (MLLMs) \cite{pattnayak2024surveylargemultimodalmodel,Bradshawjnumed.124.268072}. Complex and unbiased data that accurately reflect the richness of reality are necessary to train these models effectively. This challenge is particularly pronounced in specialized fields, such as nuclear medicine, where access to sufficient clinical data for model refinement is limited \cite{Bradshawjnumed.124.268072}. Furthermore, the need for task-specific and domain-specific datasets adds to the complexity of data collection and integration processes \cite{pattnayak2024surveylargemultimodalmodel}.

Large-scale multimodal datasets obtained from the Internet, such as image-caption pairs, often contain mismatches or irrelevant content between images and their corresponding texts. This issue arises mainly because these image-text pairs are optimized for search engines rather than for precise multimodal alignment. Consequently, models trained on such noisy data may struggle to generalize effectively. Tong et al.~\cite{tong2024eyeswideshutexploring} identifies specific instances where advanced systems like GPT-4V struggle with VQA due to inaccurate visual grounding. To address this problem, several approaches have been proposed to improve data quality.

Kim et al.~\cite{kim2024hyperbolicentailmentfiltering} introduced a novel methodology called hyperbolic entailment filtering (HYPE) , which goes beyond traditional CLIP-based filtering by incorporating both alignment and specificity metrics. HYPE leverages hyperbolic embeddings and entailment cones to filter out samples with underspecified or meaningless semantics, ensuring better cross-modal alignment and modality-wise meaningfulness.

Other notable efforts include Nguyen et al. \cite{nguyen2023improving}, who tackled noise in web-scraped datasets by using synthetic captions generated through image captioning models. By integrating synthetic descriptions with the original captions, they achieved improvements in data utility across multiple benchmark tasks, demonstrating that improved caption quality can significantly benefit model performance. Similarly, CapsFusion \cite{yu2023capsfusion} introduced a framework that leverages large language models to refine synthetic and natural captions in multimodal datasets, thus improving caption quality and sample efficiency for large-scale models. Furthermore, the LAION-5B dataset \cite{schuhmann2022laion5b} provides a large collection of CLIP-filtered image-text pairs, showing that combining high data volume with effective filtering can enhance the robustness and zero-shot capabilities of vision language models.

Despite these improvements, challenges remain in scalable data filtering and maintaining diversity. For example, DataComp \cite{gadre2023datacomp} has shown that even with effective filtering, achieving high-quality and diverse representation in large multimodal datasets is complex. It requires ongoing innovation in data pruning and quality assessment to ensure that models trained on these datasets generalize effectively across domains. HYPE's ability to consider both cross-modal alignment and intra-modal specificity offers a promising direction for addressing these limitations, especially in large-scale settings.

In summary, while synthetic captioning and large-scale filtering methods have improved the quality of multimodal datasets, further advances in scalable filtering techniques and diversity retention are needed to fully address the challenges associated with web-scraped multimodal datasets.

\subsection{Scale of Training Datasets Challenge}

Another significant challenge in multimodal learning is acquiring sufficiently large and high-quality datasets for model training, particularly for combining vision and language tasks. There is a pressing need for extensive and reliable datasets that can be used to train models effectively across a variety of tasks. For instance, the introduction of the LAION-5B dataset, comprising billions of CLIP-filtered image-text pairs, has provided a scalable, open-source dataset that supports training and fine-tuning large-scale vision-language models, helping democratize access to high-quality data \cite{schuhmann2022laion5b}. Similarly, the WIT dataset enables multimodal, multilingual learning by offering a curated, entity-rich dataset sourced from Wikipedia, featuring a high degree of concept and language diversity, which has proven beneficial for downstream retrieval tasks \cite{srinivasan2021wit}.

Although these datasets represent substantial progress, scalability and data quality remain challenging. For example, \cite{wang2023datareduction} proposes compressing vision-language pre-training (VLP) datasets to retain essential information while reducing redundancy and misalignment, resulting in a smaller but higher-quality training set. Additionally, scaling techniques like sparse mixture of experts (MoE) \cite{shen2023scaling} aim to improve the efficiency of large models by training specialized sub-models within a unified framework, balancing compute costs and performance. While these innovations are steps toward addressing data scale and quality challenges, efficient access to diverse and large datasets for multimodal learning remains a difficulty for the research community.

\subsection{Ethical Bias}

Ethical bias represents a critical and under-addressed challenge in multimodal alignment and fusion. These systems often inherit and even amplify biases present in individual modalities, such as language, vision, or speech. Studies have demonstrated that fusion processes may introduce new forms of unfairness not present in unimodal inputs, due to disparate representations and unequal weighting of modalities~\cite{yan2020mitigating,roger2023towards}. For example, \cite{yan2020mitigating} revealed that different modalities in personality assessment contribute uniquely to bias, and simple fusion can worsen demographic disparities. Similarly, \cite{zhang2023theory} developed a theoretical framework showing that multimodal networks often suffer from unimodal bias, where a dominant modality suppresses others, leading to skewed outputs. Ethical alignment frameworks, such as RoBERTa-based classifiers trained on human ethical feedback, have been proposed to better detect and mitigate bias in multimodal responses~\cite{roger2023towards}. Moreover, novel alignment datasets specifically designed to reduce gender bias in language models are also being explored to guide fairer training processes~\cite{zhang2024genderalign}. These findings underscore the need for standardized bias auditing protocols and ethically-aware training pipelines when developing multimodal systems.

\section{Discussion and Future Directions}

Multimodal alignment and fusion techniques have made significant strides in enabling complex reasoning capabilities across tasks such as visual question answering, spatial localization, and semantic composition. Recent developments demonstrate that the effectiveness of these techniques is highly task-dependent, with distinct strategies outperforming in compositional versus spatial reasoning scenarios.

For spatial reasoning, recent research underscores the continued value of incorporating explicit spatial alignment even in deep learning architectures. Wang et al.\ demonstrated that spatially aligning medical images before feeding them into deep fusion networks leads to notable improvements in diagnostic accuracy. Their addition of a spatial transformer network (STN) module further enhanced this effect, providing adaptive alignment during training \cite{wang2021role}. Similarly, the MulFS-CAP framework bypasses traditional registration by jointly learning alignment and fusion in a unified network, which proves especially effective for unregistered infrared-visible image pairs \cite{li2025mulfscap}. These results suggest that both explicit and implicit spatial alignment strategies are essential for spatially sensitive fusion tasks.

In contrast, compositional reasoning benefits more from methods emphasizing feature-level integration and cross-modality contextual alignment. ST-Align introduces a foundation model for spatial transcriptomics that aligns image and gene data across spatial scales, enabling nuanced cross-level semantic understanding \cite{lin2024stalign}. Meanwhile, Set-CLIP addresses the low-alignment data challenge by learning modality alignment through distribution-level semantic constraints, demonstrating that strong performance can be achieved even in semi-supervised settings \cite{song2024setclip}. This highlights the importance of representation space regularization and contrastive learning in achieving effective compositional fusion.

From these studies, several \textbf{generalizable lessons} emerge:
\begin{itemize}
    \item \textit{Implicit fusion architectures} with built-in alignment (e.g., STN, shared encoders) are increasingly favored for practical deployment due to reduced preprocessing demands.
    \item \textit{Cross-level fusion frameworks}, such as those combining local and global spatial cues, significantly enhance performance in hierarchical reasoning tasks.
    \item \textit{Modality-specific feature preservation}, via dictionaries or attention-based fusion, allows models to better retain complementary information during integration.
    \item \textit{Distribution-aware alignment strategies} improve robustness in data-scarce or weakly aligned settings.
\end{itemize}

Future directions include developing unified multimodal architectures that balance task-specific performance with generalization. More interpretable fusion methods are also critical, especially in clinical or scientific domains. Finally, the field would benefit from standardized multimodal benchmarks that isolate spatial and compositional reasoning tasks, enabling more consistent evaluation of fusion techniques.

Overall, effective multimodal fusion requires a deliberate choice of alignment and integration strategies, tailored to the reasoning demands of the target application. Spatial tasks benefit from precise alignment techniques, while compositional reasoning favors context-aware and semantically structured fusion methods.

\section{Conclusion}
\label{sec:conclusion}

Multimodal alignment and fusion are fundamental to advancing artificial intelligence systems capable of understanding and interacting with complex real-world data. Our survey has provided a comprehensive review of over 200 studies, categorizing existing techniques from both structural and methodological perspectives. Despite notable progress—particularly in contrastive learning, attention-based models, and LLM-driven architectures—achieving robust and scalable integration across modalities remains a significant challenge.

Current approaches continue to grapple with critical issues, including modality misalignment, inconsistent data quality, and high computational overhead. Additionally, the modality gap and the scarcity of large-scale, high-quality datasets hinder the full realization of effective multimodal learning. Although many techniques have advanced the field, limitations persist in aligning heterogeneous signals and managing data noise.

To bridge these gaps, future research should prioritize the development of adaptive, noise-resilient frameworks, capable of dynamically adjusting to diverse modalities while maintaining interpretability. Promising directions include efficient token-level fusion mechanisms, cross-modal graph reasoning, and alignment-aware training objectives that explicitly mitigate the modality gap. Furthermore, continued innovation in dataset construction, annotation quality, and filtering strategies—such as hyperbolic entailment filtering and synthetic captioning—will be essential for improving model robustness.

By addressing these challenges, the next generation of multimodal systems can become more versatile, efficient, and generalizable, paving the way for broader deployment in domains such as healthcare, autonomous systems, and human-computer interaction.

\small
\bibliographystyle{plain}
\bibliography{ref}

\end{document}